\documentclass[lettersize,journal]{IEEEtran}
\usepackage{stfloats}
\usepackage{url}
\usepackage{verbatim}

\usepackage[T1]{fontenc}
\usepackage{aecompl}

\usepackage{tabularx}
\usepackage{amsmath,amssymb,amsfonts}
\usepackage{graphicx}
\usepackage{algorithm}
\usepackage{algpseudocode}
\usepackage{subfig}
\usepackage{overpic}
\usepackage[colorlinks=true, allcolors=blue]{hyperref}
\usepackage{multirow}
\usepackage{threeparttable}
\usepackage{bm}
\usepackage{cite}
\usepackage{amsthm}
\usepackage{xcolor}

\newtheorem{Sup}{\bf Assumption}
\newtheorem{lemma}{\bf Lemma}
\newtheorem{theorem}{\bf Theorem}

\newtheorem{remark}{Remark}

\newcommand{\mG}{\mathcal{G}}

\newcommand{\mN}{\mathcal{N}}
\newcommand{\mD}{\mathcal{D}}
\newcommand{\bw}{\boldsymbol{w}}

\newcommand{\DG}{\mathcal{D}^G}
\newcommand{\bfg}{\boldsymbol{g}}
\newcommand{\bfv}{\boldsymbol{v}}
\newcommand{\bfw}{\boldsymbol{w}}
\newcommand{\bfp}{\boldsymbol{p}}

\begin{document}
	
	\title{Balanced Collaborative Exploration via Distributed Topological Graph Voronoi Partition}
	
	\author{Tianyi Ding, Ronghao Zheng$^{\dag}$, Senlin Zhang, Meiqin Liu
		
		\thanks{The authors Tianyi Ding, Ronghao Zheng, Senlin Zhang, and Meiqin Liu are with the College of Electrical Engineering, Zhejiang University, Hangzhou 310027,  China. Senlin Zhang is also with the Jinhua Institute of Zhejiang University, Jinhua 321036, China. Meiqin Liu is also with the National Key Laboratory of Human-Machine Hybrid Augmented Intelligence, Xi’an Jiaotong University, Xi’an 710049, China. All authors are also with the National Key Laboratory of Industrial Control Technology, Zhejiang University, Hangzhou 310027, China. Emails: {\tt\small \{ty\_ding, rzheng, slzhang, liumeiqin\}@zju.edu.cn}}
		
			\thanks{$^{\dag}$Corresponding author}}
	



\maketitle

\begin{abstract}
	This work addresses the collaborative multi-robot autonomous online exploration problem, particularly focusing on distributed exploration planning for dynamically balanced exploration area partition and task allocation among a team of mobile robots operating in obstacle-dense non-convex environments. 
	We present a novel topological map structure that simultaneously characterizes both spatial connectivity and global exploration completeness of the environment. The topological map is updated incrementally to utilize known spatial information for updating reachable spaces, while exploration targets are planned in a receding horizon fashion under global coverage guidance. 
	A distributed weighted topological graph Voronoi algorithm is introduced implementing balanced graph space partitions of the fused topological maps. Theoretical guarantees are provided for distributed consensus convergence and equitable graph space partitions with constant bounds.
	A local planner optimizes the visitation sequence of exploration targets within the balanced partitioned graph space to minimize travel distance, while generating safe, smooth, and dynamically feasible motion trajectories.
	Comprehensive benchmarking against state-of-the-art methods demonstrates significant improvements in exploration efficiency, completeness, and workload balance across the robot team.
\end{abstract}


\section{Introduction}
Autonomous exploration via multi-robot systems, which leverages robotic systems to map unknown environments cooperatively, is a critical capability for applications such as inspection, search-and-rescue, and disaster response \cite{2015_Bircher_ICRA}, \cite{2017_Erdlj}, \cite{2012_Marconi}. Multi-robot systems offer substantial advantages, including accelerated exploration and enhanced fault tolerance. Despite their potential, developing robust and efficient multi-robot exploration systems remains challenging due to suboptimal task allocation, and inefficient coordination strategies.

Previous collaborative exploration approaches often rely on centralized controllers \cite{Goal_assignment}, \cite{multi_dong_2019}, which are impractical in real-world scenarios with unreliable or range-limited connectivity. Decentralized coordination methods have been proposed to mitigate these issues \cite{RACER}, \cite{pairwise}, \cite{bi_cure_2024}
yet many multi-robot exploration approaches still suffer from critical inefficiencies. 
First, the exploration task load among robots is generally imbalanced in unknown environments, resulting in underutilization of some robots while others are overloaded.
Existing task allocation strategies, such as Voronoi partition-based \cite{bi_cure_2024} and pairwise interaction-based \cite{RACER}, \cite{pairwise} approaches, exhibit limitations in obstacle-dense non-convex environment and scalability. 
Second, insufficient consideration of global coverage results in myopic exploration strategies, 
where robots prioritize local frontiers without optimizing for complete environment mapping. Without a holistic exploration planner, robots may leave significant areas unexplored or inefficiently allocate resources, prolonging mission completion time.

Recently, topological maps have been widely adopted for robotic exploration and path planning \cite{MR-TopoMap}, \cite{MR-DTG}, \cite{FALCON}. As an efficient spatial representation, topological maps exhibit superior computational efficiency and communication effectiveness for planning tasks. However, their full potential in enhancing exploration efficiency remains underutilized due to several limitations.
First, the efficient topological exploration map generation remains challenging, particularly in maintaining accurate environmental connectivity representation while ensuring globally complete exploration guidance in obstacle-dense non-convex environments.
Second, current approaches lack robust discrete space decomposition algorithms for topological exploration maps capable of dynamic balancing, which is essential for effective multi-robot task allocation in unknown environments.

\begin{figure*} [t]
	\centering
	\begin{overpic}[width=18cm,height=11.2cm,trim=0 0 0 0, clip]{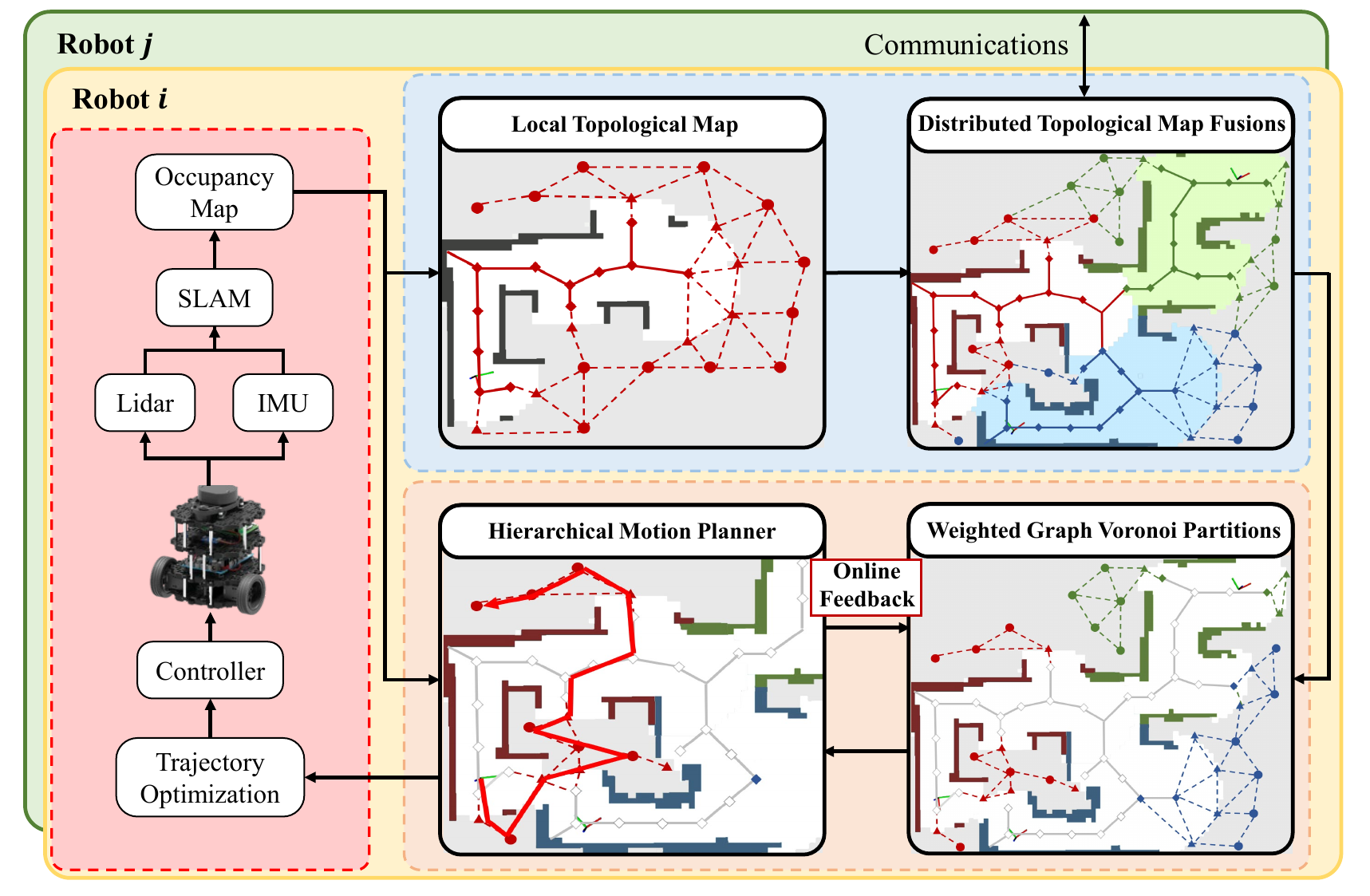}
	\end{overpic}
	\caption{Overview of the proposed balanced online collaborative exploration planner. The framework comprises two main components: constructing an efficient topological map that captures environmental connectivity and exploration completeness in obstacle-dense, non-convex environments; and implementing balanced dynamic exploration task planning with the distributed graph Voronoi guidance.}
	\label{fig:system_overview}
\end{figure*}

To overcome these limitations, we propose a balanced collaborative exploration planner via a novel distributed weighted topological graph Voronoi partition. An overview is shown in Fig.~\ref{fig:system_overview}.
To enable efficient collaborative exploration, each robot dynamically constructs a topological map based on its real-time occupancy information. Specifically, topological nodes in unknown regions are generated by computing frontiers and centroids within sliding observation windows, while the topology of known spaces is incrementally built through generalized Voronoi diagrams (GVDs). The resulting topological map simultaneously provides global exploration guidance and environmental connectivity representation. 
Furthermore, a distributed weighted graph Voronoi partition is employed to achieve balanced segmentation of the topological map. Theoretically, this segmentation is guaranteed to converge to equilibrium under distributed conditions through a max-min squeezing method. 
Subsequently, the local planner dynamically determines the visitation sequence of partitioned topological nodes through a hybrid approach combining information gain metrics with asymmetric traveling salesman problem (ATSP) formulation.
Finally, the system generates kinematically-feasible and collision-free trajectories that satisfy robot dynamic constraints. Notably, both the visitation sequence optimization and motion planning processes are accelerated through topology-guided search strategies, significantly improving computational convergence. In summary, our contributions are as follows.
\begin{enumerate}
	\item An incremental topological mapping approach in obstacle-dense environments for collaborative exploration guidance, simultaneously ensuring global exploration completeness and environmental connectivity representation.
	\item A distributed weighted topological graph Voronoi partition algorithm for balanced  exploration task allocations with theoretical guarantees for consensus and equitable non-convex space partitions.
	\item A balanced collaborative online exploration planning algorithm, which exploration task load among robots can dynamically balanced in unknown environments.
\end{enumerate}

The rest of this article is organized as follows. We review related works in Section~\ref{section:realted_work} and formulate the collaborative exploration and Voronoi partition problems in Section~\ref{section:formulation}. The distributed weighted topological graph Voronoi partition algorithm and its theoretical analysis are detailed in Section~\ref{section:graph_Voronoi}. Then, in Section~\ref{section:online_exploration}, we introduce the balancing online collaborative exploration planning algorithm. In Section~\ref{section:experiments}, we evaluate the performance of our proposed distributed multi-robot exploration in several simulation and real-world experiments. Finally, Section~\ref{section:conclusion} concludes this article.

\section{Related work}
\label{section:realted_work}
The problem of autonomous robotic exploration demonstrates significant variations in research emphasis depending on applications \cite{exploration_survey}, including rapid exploration for efficient area coverage \cite{RACER}, \cite{zhou_fuel_2021}, \cite{zhang_falcon_2025}, robust autonomous exploration under constrained localization \cite{2024_huang_localization} and communication~\cite{2024_tellaroli_iros} conditions, and task-oriented exploration prioritizing high-precision reconstruction or object detection capabilities \cite{2023_papathrodorou_icra}, \cite{2024_luo_ral}. This paper concentrates on the collaborative challenges inherent in multi-robot exploration systems. Multi-robot collaborative exploration, space partition methods and topological exploration will be discussed in this section.
\subsection{Multi-robot Exploration}
In single-robot exploration, planning focuses on generating informative viewpoints to map unknown environments and optimizing trajectories from the robot's current position \cite{zhou_fuel_2021}, \cite{zhang_falcon_2025}. For multi-robot systems, coordination extends this framework by introducing two key components: collaborative generation and fusion of shared viewpoints, and cooperative evaluation and allocation of these targets.

\textit{1) Collaborative Exploration Target Generation:} 
Existing robotic autonomous exploration methods primarily employ sampling-based and frontier-based approaches for generating exploration viewpoints. 
The classic sampling-based method~\cite{gonzalez-banos_navigation_2002} utilizes rapidly-exploring random trees (RRT) to randomly generate candidate viewpoints, subsequently selecting the next best view (NBV) through information gain optimization. 
Building upon this foundation, Bircher et al. \cite{2016_bircher_icra} introduces the NBV concept into online exploration by implementing a receding horizon scheme. 
Witting et al. \cite{2016_witting_iros} enhances the sampling process by incorporating navigation history to efficiently locate informative regions and prevent local minima entrapment. 
In contrast, frontier-based method~\cite{1997_Yammauchi_CIRA} defines candidate viewpoints as the boundaries between explored and unexplored regions within a specified exploration range. 
Recent advancements by Zhou et al. \cite{zhou_fuel_2021} further develop this approach through the frontier information structure (FIS), enabling direct target generation from frontier clusters in an incremental manner.

The aforementioned exploration target generation methods fundamentally rely on environmental map estimation. 
Distributed estimation techniques are used in multi-robot localization \cite{2014_atanasov_cdc}, multi-robot mapping \cite{2024_ding_ral}, \cite{zobeidi_dense_2022}, and multi-robot SLAM \cite{lajoie_swarm-slam_2024}, \cite{2022_tian_tro}. 
Asgharivaskasi et al. \cite{ROAM} develop a distributed Bayesian semantic mapping approach through maximization of local sensor observation log-likelihood while enforcing consensus constraints.
Ding et al. \cite{2024_ding_ral} propose distributed sparse approximation method to improve computation efficiency of observations, and further improve the communication efficiency via distributed event-triggered communication mechanism in \cite{2025_ding_ral}.
Dong et al. \cite{2022_dong_ral} introduce distributed Gaussian mixture models for environmental representation, demonstrating significantly reduced communication overhead compared to traditional uniform resolution occupancy grid mapping approaches. 

Distributed map fusion methodologies empower multi-robot systems to obtain comprehensive global information for effective exploration target generation.
However, it is important to note that the selection of map representation and fusion strategy should be guided by specific mission requirements and resource constraints. 
For example, certain cooperative localization tasks can be executed using feature maps alone \cite{2023_Stathoulopoulos_icra}.
Zhou et al. \cite{RACER} develop low-resolution heterogeneous grid maps to decompose exploration space and share between robots to generate frontiers. 
Moreover, topological maps are developed to replace the dense occupancy maps for communication restricted environment in \cite{yang_active_2024}, \cite{MR-TopoMap} and \cite{MR-DTG}. 
In this work, we employ a hybrid topological map structure that simultaneously characterizes environmental connectivity and exploration completeness, while enabling lightweight inter-robot communication and topological map fusion.

\textit{2) Collaborative Exploration Target Allocation:} 
The allocation of generated exploration targets to individual robots follows two important metrics: one is \textit{updated information}, quantifying the map information gain achievable from the generated targets; the other is \textit{exploration completeness}, representing the achievable environmental coverage through target visitation. As exploration progresses, target generation and task allocation are updated dynamically. The primary challenge lies in achieving distributed maximization of both global \textit{updated information} and \textit{exploration completeness} during dynamic processes.

Distributed optimization methods are prevalent in exploration target allocations. The algorithms introduced in \cite{2014_Gharesifard_tac} and \cite{2015_Nedic_tac} provide frameworks for decentralized gradient-based optimization in Euclidean space while accommodating various constraints including time-varying conditions and network communication asymmetries.  Informative path planning (IPP) methods are developed to maximize information gains through explorations, such as mutual information \cite{2014_geoffrey_ijrr}, \cite{2024_Jakkala_icra} and map uncertainty \cite{2024_Newaz_ral}. 
Ergodic metric \cite{2025_sun_tro}, \cite{seewald_energy-aware_2024} is proposed to enable optimal exploration of an information distribution with guaranteed asymptotic coverage of the search space.
Asgharivaskasi et al. \cite{ROAM} propose a decentralized Riemannian optimization that 
the objective functions combined with the collision, information gain and aggregate distance scores can converge to a local optimum of the consensus constraint. 
Wu et al. \cite{BGE} introduce a Bayesian-guided evolutionary strategy to predict information gain, cost, and repulsive potential field for the robot-to-frontier assignments.
Despite these advancements, these approaches typically evaluate the travel costs of robots in Euclidean space, which fails to accurately reflect the actual travel cost in obstacle-rich environments and cause unreasonable target allocations. Moreover, these methods often yield greedy suboptimal solutions based on current state estimations in online exploration scenarios, potentially leading to myopic target allocation strategies.

To enable non-myopic exploration target allocation, Faigl et al. \cite{2012_Faigl_iros} formulate the multi-robot exploration task allocation as a multiple traveling salesman problem (MTSP), where targets are clustered via K-means and allocated according to TSP distance metrics. Zhou et al. \cite{RACER} extend this framework through a pairwise interaction method incorporating both distance costs and capacity constraints under a capacitated vehicle routing problem (CVRP) formulation, simultaneously minimizing total coverage path length while enforcing local path length constraints. TSP-based methods are capable of allocating multiple targets to robots with the shortest or near-shortest coverage paths. However, TSP is a NP-hard problem and the computational complexity increases dramatically with the number of targets, necessitating the clustering approach adopted in \cite{2012_Faigl_iros} and restricting optimization to pairwise allocations in \cite{RACER}.
\subsection{Space Partitions for Multi-robot Systems}
Space partition methods decompose the exploration area into subregions based on specific criteria, employing techniques such as Voronoi diagrams \cite{Voronoi_survey} or clustering algorithms~\cite{clustering_survey}. 
Notably, the generation of exploration targets and space partition can proceed independently. 
While some approaches \cite{2012_Faigl_iros} initially generate targets before applying space partitions, 
alternative methods \cite{bi_cure_2024}, \cite{RACER} first partition the environment into subregions before target generation within each allocated area.

Extensive research has been conducted on convex space partition and coverage problems, including balanced Voronoi partitions~\cite{2010_Cortes_tac}, \cite{pavone_distributed_2011}, perception-constrained space partitions~\cite{boardman_limited_2017}, and heterogeneous swarm space partitions~\cite{2024_zhang_acc}.
However, in non-convex environments with dense obstacles, achieving efficient balanced spatial decomposition remains challenging. 
Seoung et al. \cite{2016_Seoung_ijrr} develop a structured triangulation method for non-convex environment coverage in exploration tasks, 
while Durham et al. \cite{2012_Durham_tro} introduce a pairwise partition rule to dynamically update territory ownership in discrete grid representations. 
Bhattacharya et al. \cite{bhattacharya_multi-robot_2014} propose a wavefront-based Voronoi partition approach where collision points between wavefronts in discrete grids define Voronoi tessellation boundaries. 
Dong et al. \cite{MR-DTG} design a multi-robot dynamic topological map, which is generated by grid-level Dijkstra search and achieve graph space partition via graph Voronoi \cite{GraphVor}.
Although these discrete space search-based partition techniques provide systematic solutions, they typically require substantial computational resources for online implementation in obstacles-dense environments.

\subsection{Exploration with Topological Map Guidance}
Topological maps provide an efficient sparse environmental representation that can be utilized across various stages of robotic autonomous explorations, such as communication restricted coordination \cite{MR-TopoMap}, relative locations \cite{2024_Bai_iros}, exploration planning \cite{FALCON}, \cite{vutetakis_active_2024}, and path planning \cite{de_groot_topology-driven_2025}.
Here we focus on efficient exploration and path planning.
To implement non-myopic exploration planning, Zhang et al. \cite{FALCON} develop a coverage path guidance single-robot exploration planner, which maintains a dynamic connectivity graph via restricted A* searches. 
Vutetakis et al. \cite{vutetakis_active_2024} propose a sampled-based topological roadmap to store environment reachability for guiding view sampling and refinement to ensure maximum coverage of the unmapped space.
In \cite{de_groot_topology-driven_2025}, a global planner iteratively generates trajectories in distinct homotopy classes by visibility probabilistic roadmaps, which guides a local planner as initial guess and through a set of homotopy constraints.
The generalized Voronoi diagrams (GVDs) consist of all points that are equidistant to multiple obstacles \cite{GVD_IJRR}, \cite{GVD_TRO}, \cite{2009_Nidhi_ras}.  
Wen et al. \cite{wen_gvd_2025} develop a GVD-based topological map and topological-level A* methods to reduces the path search complexity.
Dong et al. \cite{MR-DTG} design a dynamic topological map to store explored nodes and achieve graph space partition via graph Voronoi to prevent exploring the same area concurrently. 
These methods leverage topological maps to efficiently represent explored environmental connectivity and foresighted global coverage information, significantly enhancing both exploration and path planning efficiency. However, the online maintenance of such representations through frequent search \cite{MR-DTG}, \cite{FALCON} or sampling \cite{vutetakis_active_2024}, \cite{de_groot_topology-driven_2025} operations incurs substantial computational overhead, with the topological map fusion process in multi-robot systems requiring additional node computations due to inconsistent sampling across agents.

In this work, we introduce a novel hybrid topological map that distinctively integrates global coverage guidance with environmental connectivity representation, contrasting with prior approaches in \cite{MR-DTG}, \cite{FALCON}, \cite{vutetakis_active_2024}. Our framework employs incremental generation methods with environment consistency, ensuring computationally efficient online representation and distributed map fusion. Furthermore, advancing beyond \cite{MR-DTG}, \cite{GraphVor}, we develop generalized weighted graph Voronoi partitions coupled with distributed weighted iteration methods, providing balanced exploration task allocation with theoretical guarantees for both consensus achievement and equitable non-convex space partitions.

\section{Problem formulation and Preliminaries}
\label{section:formulation}
\subsection{Problem Formulation}
Consider a team of robots $V=\{1,...,i,...,n\}$. Collaborative exploration seeks to construct a complete map $\mathcal{M}$ of a unknown environment.
Each robot performs distributed state estimation to localize itself $\mathbf{r}_i$ while simultaneously building local maps $\mathcal{M}_i$ using onboard sensors. Each robot exchanges map information with neighbors when communication is available and update distributed fused map. Through an iterative process, each robot continuously evaluates the current distributed map state to determine exploration targets in a distributed manner, then employs local motion planning to navigate safely. This cycle of information gathering, communication, task allocation, and navigation persists until the collective effort yields a globally consistent environmental map, i.e., $\mathcal{M}_i \rightarrow \mathcal{M},~i \in V$.

This paper focuses on the balanced collaborative exploration problem with dynamic exploration task allocation, particularly in obstacle-dense, non-convex environments.
Consider a exploration state-space $\mathcal{X} \in \mathbb{R}^d$, where $\mathcal{X}_\mathrm{obs}$ represents the obstacle space and $\mathcal{X}_\mathrm{free}=\mathcal{X}\setminus\mathcal{X}_\mathrm{obs}$ denotes the obstacle-free space.
The length of the path $\mathcal{P}_i$ for robot $i$ can be computed as
\begin{equation*}
	\mD^E(\mathcal{P}_i) = \sum_{\mathbf{x}_{m},\mathbf{x}_{m+1}  \in \mathcal{P}_i} \mD^E(\mathbf{x}_{m},\mathbf{x}_{m+1}),
\end{equation*}
where $\mathcal{D}^E$ denotes the Euclidean distance in state-space $\mathbf{x}_m \in \mathcal{X}$.
Then the balanced collaborative exploration problem can be described as:

\textit{Problem 1 (Balanced Collaborative Exploration):} Given a set of exploration targets $N$, a team of robots $V$ and the balanced threshold $\hat{B}$, 
the balanced collaborative exploration problem is aimed to 
find a subset of exploration targets $N_i$ and a exploration path $\mathcal{P}_i$ for each robot $i$
such that the exploration paths of all robots visiting their allocated exploration targets minimize the total travel distance while satisfying the exploration workload balance and safety constraint, 
\begin{equation*}
	\begin{aligned}
		\min_{\{\mathcal{P}_i\}} & \sum_{i \in V} \mD^E(\mathcal{P}_i), \\
		\text{s.t.}&~ N_i \subseteq \mathcal{P}_i \subset \mathcal{X}_\mathrm{free}, \\
		&~ N = \bigcup_{i\in V} N_i, \\
		&~ \lvert \mD^E(\mathcal{P}_i) - \mD^E(\mathcal{P}_j) \lvert \leq \hat{B},~\forall i, j \in V.
	\end{aligned} 
\end{equation*}

The constraint $N_i \subset \mathcal{P}_i \subset \mathcal{X}_\mathrm{free}$ in Problem~1 ensures that each robot's trajectory $\mathcal{P}_i$ remains collision-free while visiting all assigned targets in $N_i$. And the constraint $N = \bigcup_{i\in V} N_i$ ensures that all exploration targets are assigned and visited. The final constraint enforces balanced exploration path lengths across all robots.

To provide context for the proposed method, we first present preliminaries related to generalized Voronoi diagrams (GVD) and graph Voronoi partitions.

\begin{figure*} [t]
	\centering
	\begin{overpic}[width=16.2cm,height=5.cm,trim=0 0 0 0, clip]{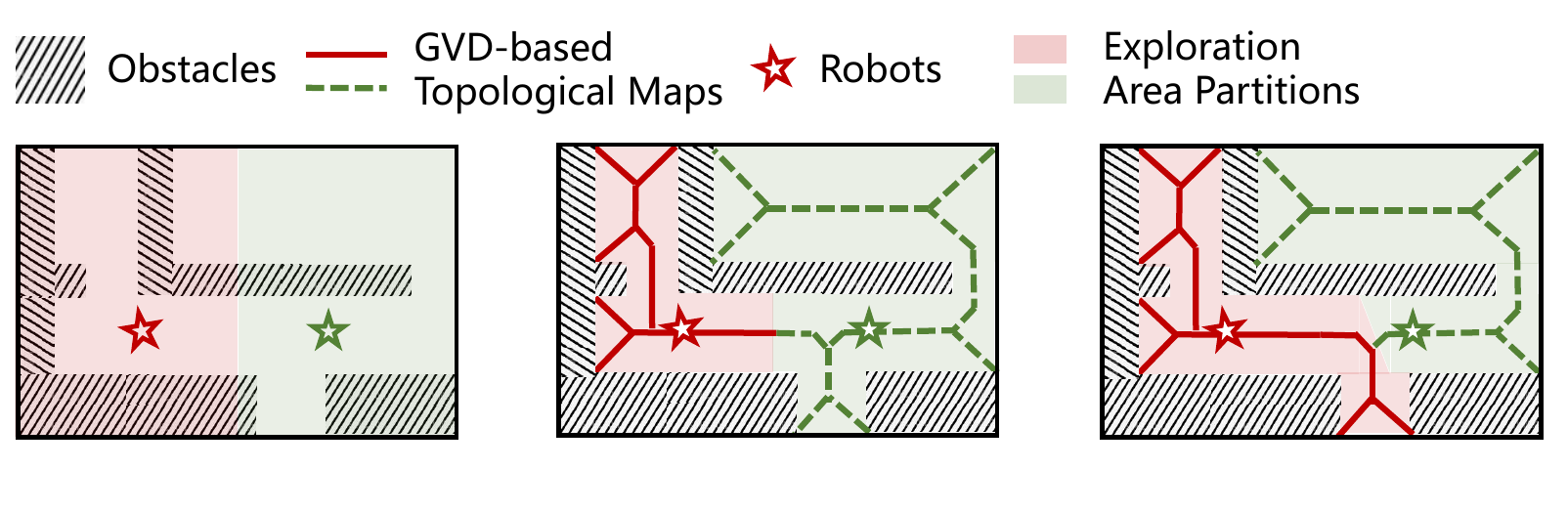}
		\put(2,1){\color[rgb]{0,0,0}{(a) Euclidean Space Voronoi}} 
		\put(41,1){\color[rgb]{0,0,0}{(b) Graph Voronoi}} 
		\put(72,1){\color[rgb]{0,0,0}{(c) Weighted Graph Voronoi}}
	\end{overpic}
	\caption{Different Voronoi partitions in the non-convex environments.}
	\label{fig:different_voronoi}
\end{figure*}

\subsection{Generalized Voronoi Diagrams}
The generalized Voronoi diagram  consists of all points that are equidistant to multiple obstacles \cite{GVD_IJRR}, \cite{GVD_TRO}, \cite{2009_Nidhi_ras}. The GVD can be formally defined as:
\begin{multline}
	\label{eq:GVD}
	\text{CF}_{ij} = \{ q\in \mathcal{X}_\mathrm{free}| 0 < \mathcal{D}^E(q,\mathcal{O}_i) = \\ \mathcal{D}^E(q,\mathcal{O}_j) < \hat{D}, \forall \mathcal{O}_i,~\mathcal{O}_j \in \mathcal{X}_\mathrm{obs}\}, 
\end{multline}

\begin{equation*}
	\text{GVD} = \bigcup_{i,j} \text{CF}_{ij},
\end{equation*}
where the distance $\mathcal{D}^E(q,\mathcal{O}_i)$ denotes the minimum Euclidean distance between the point $q\in \mathcal{X}_\mathrm{free}$ and the point in the obstacle $\mathcal{O}_i$. 
The parameter $\hat{D}$ represents the maximum search range.
The GVDs provide an efficient representation of environmental connectivity. The generation of GVD is introduced in Section~\ref{incremental_topo}.

\subsection{Graph Voronoi Partitions}
Graph Voronoi partitions \cite{GraphVor} decompose the graph space into disjoint regions based on proximity to designated central nodes. 
Considering an undirected graph $G=(N,E)$, define the distance $\DG$ in the graph space.
Denote $\DG(e) : E \rightarrow \mathbb{R}^+$ as the edge length for a edge $e \in E$.
For nodes $\bfv_q,~\bfv_p \in N$, a shortest path, denoted $\mathcal{P}(\bfv_q,\bfv_p) = \{e_1, ..., e_k\}$, is a sequence of edges with the smallest sum of edge lengths, where $e_m=(\bfv_{m-1}, \bfv_{m}) \in E$, with $\bfv_0 = \bfv_q,~\bfv_k = \bfv_p$. Then, define
the distance $\DG(\bfv_q,\bfv_p)$ between nodes $\bfv_q,~\bfv_p \in N$ in the graph $G$ as 
\begin{equation*}
	\DG(\bfv_q,\bfv_p) = \sum_{e_m \in \mathcal{P}(\bfv_q,\bfv_p) } \DG(e_m).
\end{equation*}

The graph Voronoi partition $\{N_1,...,N_n\}$ for a set of central nodes $\mG=\{\bfg_1,...,\bfg_n\}\subseteq N$ satisfies that for each node $\bfv\in N_i$,~$\DG(\bfg_i,\bfv)\leq \DG(\bfg_j, \bfv)$ for all $j \in \{1,...,n\}$.
Fig.~\ref{fig:different_voronoi}(b) illustrates an example of a graph Voronoi partition.

\section{Distributed Topological Graph Voronoi Partitions}
\label{section:graph_Voronoi}
Balanced exploration workload is crucial for ensuring efficiency, scalability, and optimal energy utilization in multi-robot collaborative exploration systems.
In this section, we focus on balanced partition of non-convex exploration spaces within offline known environment. 
The online collaborative exploration will be discussed later in Section~\ref{section:online_exploration}.

In this section, environmental topological map is constructed using GVD \eqref{eq:GVD}, which maps
the non-convex exploration spaces into graph space. 
A balanced graph space partition problem is considered as a simplified version of the Problem~1.

\textit{Problem 2 (Balanced Graph Space Partition):} Given a team of robots $V$ and a set of exploration targets $N$ in the non-convex exploration spaces $\mathcal{X}$, 
the balanced graph space partition problem is aimed to 
find a subset of exploration targets $N_i$ for each robot $i$
such that minimizes workload disparity in graph space, 
\begin{equation*}
	\begin{aligned}
		\min_{\{N_i\}} & \sum_{i,j \in V} \big\lvert \mD^G(N_i) - \mD^G(N_j) \big\lvert, \\
		\text{s.t.}&~ N_i \subset \mathcal{X}_\mathrm{free}, \\
		&~ N = \bigcup_{i\in V} N_i,
	\end{aligned} 
\end{equation*}
where $\mD^G(N_i)$ denotes the cumulative graph distance metric for exploration target sets $N_i$.

\subsection{Weighted Graph Voronoi Partitions}

\begin{algorithm} [t]
	\caption{Distributed Weighted Graph Voronoi Partitions for Robot $i$} 
	\begin{algorithmic}[1]
		\State Initialize $\bw_{i} = \mathbf{0}$
		\While{exist $j'$ such that $\lvert \Delta_\lambda(j',i,k) \lvert \geq B_\lambda$}
		\State Compute the load score $\lambda\big(\bfg_i,N_i(k)\big)$ by \eqref{eq:load_metric}
		\For{each $j \in \mN_i$}
		\State Communicate and share local topological maps $G_i$, power points $\mG_{W_i}$ and load scores $\lambda\big(\bfg_i,N_i(k)\big)$ with the neighbor $j$
		\State Receive $G_j,~\mG_{W_j},~\lambda\big(\bfg_j,N_j(k)\big)$ from $j$
		\If {$\lvert \Delta_\lambda(j,i,k) \lvert \geq B_\lambda$}
		\State Update the weighted $w_{ij}(k+1)$ by \eqref{eq:weight_update}
		\EndIf
		\EndFor
		\State Update the weighted graph Voronoi partitions $N_i(k+1)$ by \eqref{eq:graphVor}
		\State $k \leftarrow k+1$
		\EndWhile
	\end{algorithmic}
	\label{al:iter_graphVor_1}
\end{algorithm}

While classical graph Voronoi partition \cite{GraphVor} provides decomposition of discrete graph structures, it inherently lacks load-balancing guarantees. The workload distribution across partitioned regions depends entirely on the location of central nodes $\mG$. To address this limitation, a novel weighted graph Voronoi partition is proposed and defined as follows.

\textit{Weighted Graph Voronoi Partition:} Consider an undirected graph $G=(N,E)$ and a set of distinct power points $\mG_W=\{(\bfg_1,\bw_1),...,(\bfg_n,\bw_n)\}$, where each weight vector $\bw_i=[w_{i1},...,w_{in}]$ represents the relative weights between the $i$-th partition and others. The weighted graph Voronoi partition $\{N_1,...,N_n\}$ satisfies the following conditions:
\begin{itemize}
	\item For each node $\bfv\in N_i$,
	\begin{equation}
		\label{WGV}
		\DG(\bfg_i, \bfv)-w_{ij}\leq \DG(\bfg_j, \bfv)
	\end{equation}
	holds for all $(\bfg_j,\bw_j) \in \mG_W$.
	\item For any distinct power points, $w_{ij} < \DG(\bfg_i, \bfg_j)$.
\end{itemize}

The weight vectors $\bw_i$ enables dynamic boundary adjustment between partitions, thereby achieving partition balancing. The comparison of the different Voronoi partitions is shown in Fig.~\ref{fig:different_voronoi}.

Furthermore, we present a graph-based evaluation metric to quantify robot exploration workload distribution across current partitions. Building upon this metric, a distributed iterative weight vector update algorithm is developed that guarantees convergence to balanced workload distribution among multiple robots.

\textit{Graph-based Exploration Load Metric:} Consider a team of robots $V$ engaged in cooperative exploration, 
where a topological map $G=(N,E)$ of the environment has been constructed and partitioned into $n$ distinct sub-map $G_i=(N_i,E_i)$ assigned to robot $i$, which satisfies $N = \cup_{i\in V} N_i$ and $E = \cup_{i\in V} E_i$. 
The node position closest to the $i$-th robot location is denoted as $\bfg_i$.
Define $\mathcal{E}(\bfg_i,N_i)$ as the union of all edges along the shortest paths from a central node $\bfg_i$ to all other nodes $\bfv$ in the sub-graph $G_i$,
\begin{equation*}
	\mathcal{E}(\bfg_i,N_i) = \bigcup_{\bfv \in N_i} \mathcal{P}(\bfg_i, \bfv).
\end{equation*}
Then, the graph-based exploration load metric $\lambda(\bfg_i,N_i)$ is defined as:
\begin{equation}
	\label{eq:load_metric}
	\lambda(\bfg_i,N_i) = \sum_{e \in \mathcal{E}(\bfg_i,N_i)} \DG(e).
\end{equation}

\begin{figure} [t]
	\centering
	\begin{overpic}[width=9cm,height=4.5cm,trim=0 0 0 0, clip]{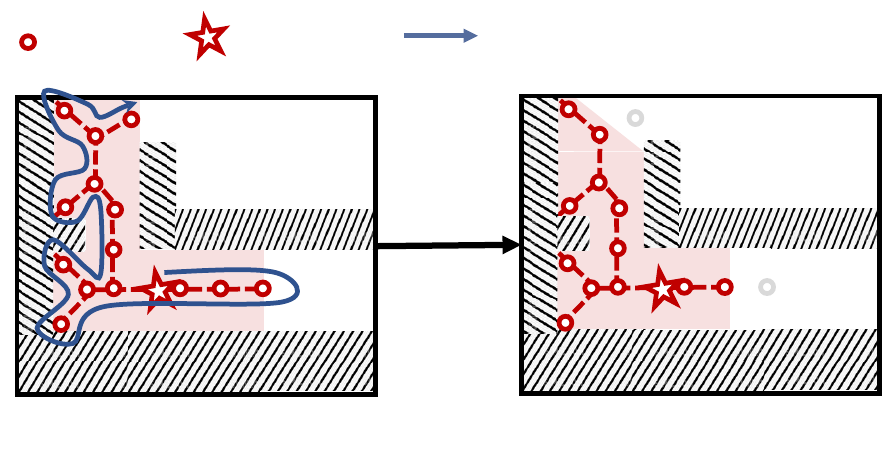}
		\put(8,44){\color[rgb]{0,0,0}{Nodes}}
		\put(30,44){\color[rgb]{0,0,0}{Robots}} 
		\put(55.5,44.5){\color[rgb]{0,0,0}{Graph Traversal Traj.}}
		
		\put(73,37){\color[rgb]{0,0,0}{$\bfv_p$}}
		\put(88,16){\color[rgb]{0,0,0}{$\bfv_q$}}
		
		\put(44,30){\color[rgb]{0,0,0}{\small lose leaf}}
		\put(45.5,26.5){\color[rgb]{0,0,0}{\small nodes}}
		
		\put(23,0){\color[rgb]{0,0,0}{$\lambda(\bfg_i,N_i) \geq \lambda(\bfg_i,N_i\setminus \{\bfv_p, \bfv_q\})$}}
	\end{overpic}
	\caption{An example of graph-based exploration load metric and its decrease property. The metric formulation guarantees an upper bound on feasible exploration paths, $\mD^E(\mathcal{P}_i) \leq 2 \cdot \lambda(\bfg_i, N_i)$.}
	\label{fig:load_metric}
\end{figure}

\begin{algorithm} [ht]
	\caption{Parallel Dijkstra for Robot $i$} \label{alg:D}
	\begin{algorithmic}[1]
		\Require Graph $G_{ci} = G_i\bigcup_{j \in \mN_i} G_j = (N_{ci},~E_{ci})$,~central nodes $\bfg_{ci} = \{\bfg_i\}\bigcup_{j \in \mN_i}\{\bfg_j\}$, weight matrix $\mathbf{W}_{ci}=[\bw_i, \bw_j]_{j \in \mN_i}$
		\State \# initialize min-heap $h$
		\For{$\bfv \in N_{ci}$}
		\If{$\bfv \in \bfg_{ci}$}
		\State $d(\bfv):=0$;~$\text{Vor}(\bfv) = \bfv$; $\lambda\big(\text{Vor}(\bfv)\big) = 0$ \State $\textbf{insert}(\bfv,h)$
		\Else
		\State  $d(\bfv):=\infty$;~$\text{Vor}(\bfv) = \text{None}$
		\EndIf
		\EndFor
		\State 
		\While {$h$ is not empty}
		\State $\bfv=\textbf{pop}(h)$
		\State mark $\bfv$
		\For{edge $E_{ci}(\bfv,\bfw)$ exists and $w$ not marked}
		\State $\Delta:= d(\bfv)+\mathcal{D}^{G_{ci}}(\bfv,\bfw)$
		\If{$d(\bfw) = \infty$}
		\State $d(\bfw)=\Delta$;~$\text{Vor}(\bfw) = \text{Vor}(\bfv)$ 
		\State $\lambda\big(\text{Vor}(\bfv)\big) += \mathcal{D}^{G_{ci}}(\bfv,\bfw)$
		\State $\textbf{insert}(\bfw,h)$
		\ElsIf{$\Delta - \mathbf{W}_{ci}\Big(\text{Vor}(\bfw),\text{Vor}(\bfv)\Big) < d(\bfw)$}
		\State $\lambda\big(\text{Vor}(\bfw)\big) -= \mathcal{D}^{G_{ci}}(\bfw',\bfw)$
		\State $\text{Vor}(\bfw)=\text{Vor}(\bfv)$;~$\lambda\big(\text{Vor}(\bfv)\big) += \mathcal{D}^{G_{ci}}(\bfv,\bfw)$
		\State $\textbf{decrease}(d(\bfw)-\Delta,\bfw,h)$
		\EndIf
		\EndFor
		\EndWhile
		\State \text{Return $\text{Vor}(N_{ci})$, $\lambda$}
	\end{algorithmic}
	\label{al:Dijkstra}
\end{algorithm}

The exploration load is defined as the total length of these shortest-path edges removing the repeated edges. This formulation guarantees an upper bound with a constant coefficient on feasible exploration paths, which represents that at least one feasible path $\mathcal{P}(N_i)$ can be found for visiting all exploration targets in $N_i$, and satisfies $\mD^E(\mathcal{P}_i) \leq 2 \cdot \lambda(\bfg_i,N_i)$. 
Moreover, these shortest-path edges can be obtained through parallel Dijkstra's algorithm, which inherently produces them as intermediate variables during the graph Voronoi partition process. Consequently, employing this metric \eqref{eq:load_metric} for weighted graph Voronoi iterations incurs no additional computations. The implementation details of this process will be elaborated latter.

Furthermore, the metric \eqref{eq:load_metric} possesses an important property: its value strictly decreases when a leaf node rooted at $\bfg_i$ (the outermost node in the graph $G_i$) is removed from a partition $N_i$, i.e.,
\begin{equation}
	\label{eq:leaf_decrease}
	\lambda(\bfg_i,N_i) \geq \lambda(\bfg_i,N_i\setminus \{\bfv^l\}),~\forall \bfv^l \in leaf(N_i).
\end{equation}
This property matters for the convergence of the following distributed weight vector iterations. An example of the proposed load metric is shown in Fig.~\ref{fig:load_metric}.

\begin{remark}
	The proposed metric \eqref{eq:load_metric} does not strictly equal the actual motion path length required by robots, as their local motion planners further optimize the visitation sequence of assigned exploration targets, similar to the asymmetric traveling salesman problem (ATSP) in Ref.\cite{RACER}. However, since solving the ATSP is NP-hard, its prohibitive computational complexity makes it unsuitable as an exploration load metric. In contrast, the designed metric \eqref{eq:load_metric} guarantees an upper bound on feasible path lengths, and optimizing the metric \eqref{eq:load_metric} effectively reduces the actual motion path length. More results will be discussed in Section.~\ref{section: motion_planner}.
\end{remark}

\textit{Weighted Graph Voronoi Iteration}: 
Let $\mN_i$ denote the set of neighboring robots for robot $i$, where neighbors are defined as robots that share overlapping exploration regions and remain within communication range.
The distributed iteration method for computing weighted graph Voronoi partitions is presented in Algorithm~\ref{al:iter_graphVor_1}. Within the distributed communication framework, each robot $i$ acquires local topological maps $G_i$, power points $\mG_{W_i}$ and load scores $\lambda\big(\bfg_i,N_i(k)\big)$ at time $k$ from its neighboring robots $j \in \mN_i$. When exploration load imbalance is detected (line~7), the relative weights $w_{ij}$ are updated (line~8) and the graph Voronoi partitions $N_i(k)$ are recomputed (line~11) until the convergence and balance are achieved.

Following the approach in \cite{GraphVor}, the parallel Dijkstra algorithm (Algorithm~\ref{al:Dijkstra}) is used to compute the graph Voronoi partitions for each robot as
\begin{equation}
	\label{eq:graphVor}
	\begin{aligned}
		N_i = \textbf{graphVor}\big(G_i\bigcup_{j \in \mN_i} G_j,\{\bfg_i\}\bigcup_{j \in \mN_i}\{\bfg_j\}, [\bw_i, \bw_j]_{j \in \mN_i}\big),
	\end{aligned}
\end{equation}
where $G_i\bigcup_{j \in \mN_i} G_j$ denotes the merge of topological maps. The topological map merge process follows a similar approach to \cite{MR-TopoMap}, involving computation of geometric transformations between maps, merging of proximate nodes, and enhancement of edge connectivity. In Algorithm~\ref{al:Dijkstra}, the \textbf{insert}$(\bfv,h)$ operations (line~$5$ and line~$19$) insert the node $\bfv$ into the min-heap $h$ with the value $d(\bfv)$. The \textbf{decrease}$(d(\bfw)-\Delta,\bfw,h)$ operations (line~$23$) decreases the value for $w$ in min-heap $h$ by $d(\bfw)-\Delta$. 

Note that during the execution of Algorithm~\ref{al:Dijkstra}, the shortest paths from all nodes to their assigned Voronoi centroids are inherently determined. The exploration metric $\lambda$ \eqref{eq:load_metric} mentioned above can be obtained simply by accumulating or modifying the corresponding edge lengths $\mathcal{D}^{G_{ci}}$ during the process of establishing (line~$4$ and line~$18$) or altering (line~$21-22$) Voronoi partitions, without requiring additional computations. The $\mathbf{w}'$ in line~$21$ denotes the parent node from the last Voronoi center for the node $\mathbf{w}$.

Moreover,
the weight vector update for robot $i$ at time $k$ are designed as follows:
\begin{equation}
	\begin{aligned}
		\label{eq:weight_update}
		\Delta_\lambda(j,i,k) &= \lambda\big(\bfg_j,N_j(k)\big) - \lambda\big(\bfg_i,N_i(k)\big), \\
		\Delta w_{ij}(k+1) &= \Bigg\{ \begin{aligned}
			& 0, \text{if $\lvert \Delta_\lambda(j,i,k) \lvert \leq B_\lambda$}, \\
			& \gamma \cdot \mathrm{sign}\big(\Delta_\lambda(j,i,k)\big), \text{otherwise},
		\end{aligned}\\
		w_{ij}(k+1) &= w_{ij}(k) + \Delta w_{ij}(k+1),~j \in \mN_i,
	\end{aligned}
\end{equation}
where $\gamma > 0$ is a constant parameter controlling the weight update step size and $B_\lambda $ denotes the tolerance threshold for exploration load balancing. Each robot $i$ obtains the relative deviation of its current exploration load from neighboring robots $j$ via communication, and accordingly updates the graph Voronoi weights $w_{ij}$. For robot $j$ with excessive exploration load (relative to robot $i$), its weight $w_{ij}$ is increased to reduce its task allocation in subsequent partitions, thereby achieving dynamic load balancing. The design of parameters $\gamma$ and $B_\lambda $, along with convergence analysis of the exploration load, will be detailed in Section~\ref{section:graph_vor_ana_1}.

\subsection{Theoretical Analysis for Algorithm~\ref{al:iter_graphVor_1}}
\label{section:graph_vor_ana_1}

\begin{Sup}
	\label{sup:iter_bound}
	Suppose there exists a constant $\mathcal{B} > 0$ for graph $G$ such that for all $i \in V,~\bfg_i \in \mG_W$, the condition $\lvert \lambda\big(\bfg_i,N_i(k+1)\big) - \lambda\big(\bfg_i,N_i(k)\big)  \lvert \leq \gamma\mathcal{B}$ holds.
\end{Sup}

Assumption~\ref{sup:iter_bound} supposes that there exists an upper bound $\gamma\mathcal{B}$ on the load variation induced by a single weight iteration \eqref{eq:weight_update} and graph Voronoi repartition \eqref{eq:graphVor} for all robots. This assumption is justified as the exploration load \eqref{eq:load_metric} constitutes a bounded discrete topological distance.

\begin{Sup}
	\label{sup:communication}
	Communication graph between all robots is connected. For any robot $i$ and $j$, there exist a communication path.
\end{Sup}

Assumption~\ref{sup:communication} supposes that communication for robots maintain either single-hop or arbitrary multi-hop connectivity. In practical applications, however, strict adherence to Assumption~\ref{sup:communication} is not required, as robots may operate in multiple locally connected subgroups. Theorem~\ref{theo:1} subsequently guarantees that all communication-connected robots can achieve balanced exploration workload distribution through Algorithm~\ref{al:iter_graphVor_1}.

Then, we employ a max-min squeezing method to prove that Algorithm~\ref{al:iter_graphVor_1} guarantees convergence of all robots' exploration loads to an bounded equilibrium state.
Define
\begin{equation*}
	\begin{aligned}
		M_\lambda(k) &= \max_{p \in V} \lambda\big(\bfg_p,N_p(k)\big) , \\
		m_\lambda(k) &= \min_{p \in V} \lambda\big(\bfg_p,N_p(k)\big), \\
		D_\lambda(k) &= M_\lambda(k) - m_\lambda(k),
	\end{aligned}
\end{equation*}
where $M_\lambda(k)$, $m_\lambda(k)$ are the maximum and minimum exploration load scores for all robots, respectively, and $D_\lambda(k)$ is the difference.
Denote that $\bar{p}(k) \triangleq \mathop{\arg\max}_{p \in V}\lambda\big(\bfg_p,N_p(k)\big)$ and $\underline{p}(k) \triangleq \mathop{\arg\min}_{p \in V}\lambda\big(\bfg_p,N_p(k)\big)$ are the robots with the maximum and minimum load scores, respectively.

\begin{lemma}
	\label{fix_dec_D}
	Consider the weighted graph Voronoi partitions $N_i$ being updated according to Algorithm~1. Suppose that Assumption~\ref{sup:iter_bound} and \ref{sup:communication} holds. Let the parameter satisfying $B_\lambda  \geq 2\gamma \mathcal{B}$. Then, there exist constants $\kappa \in  \mathbb{N}^+$ and $\Gamma \in \mathbb{R}^+$ such that the following inequality holds:
	\begin{equation}
		\label{fix_dec_guarantee_2}
		D_\lambda(k+\kappa) < D_\lambda(k) - \Gamma,
	\end{equation}
	before the termination condition is met for all robots at time $\hat{k}$, i.e., $\lvert \Delta_\lambda(j,i,k) \lvert < B_\lambda,~\forall  k \geq \hat{k} ,~\forall i \in V,~\forall j \in \mN_i$.
\end{lemma}

\begin{proof}
	The proof of Lemma~1 is shown in Appendix A.
\end{proof}

Lemma~1 shows that the difference metric $D_\lambda$ is asymptotically decreasing during the iterations of Algorithm~1. Although the gradient magnitude and step size of this descent are non-constant due to the discrete nature of the graph space, there exists a constant lower bound for the decrement. This property consequently guarantees the convergence of Algorithm~1 in Theorem~1.

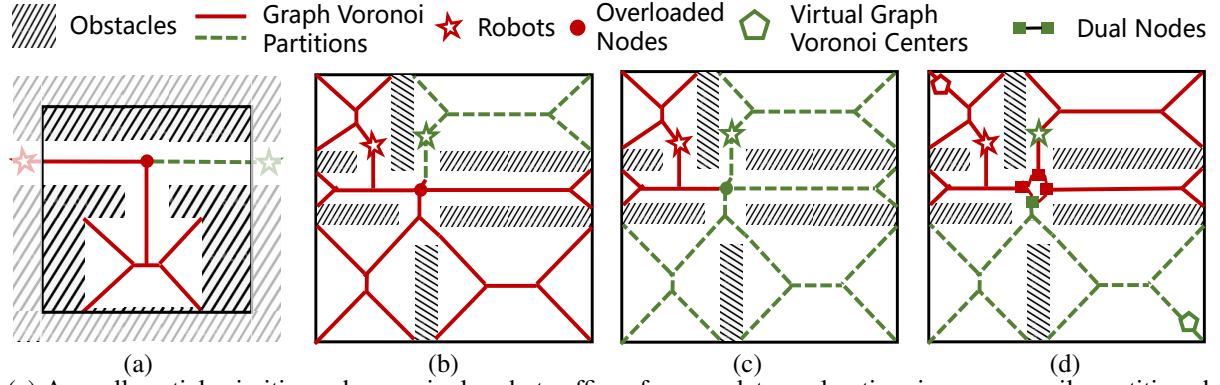
\begin{figure*} [t]
	\centering
	\begin{overpic}[width=16.2cm,height=5.cm,trim=0 0 0 0, clip]{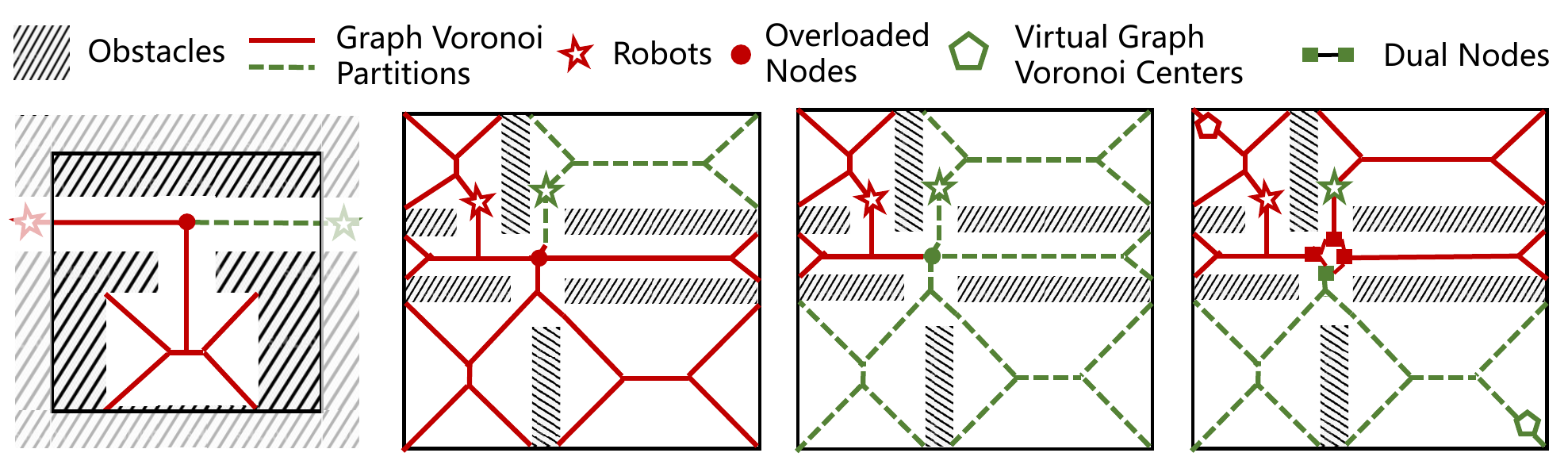}
		\put(10,-1){\color[rgb]{0,0,0}{(a)}} 
		\put(35,-1){\color[rgb]{0,0,0}{(b)}} 
		\put(60,-1){\color[rgb]{0,0,0}{(c)}}
		\put(86,-1){\color[rgb]{0,0,0}{(d)}}  
	\end{overpic}
	\caption{(a) A small spatial primitive, where a single robot suffices for complete exploration, is unnecessarily partitioned. (b)-(c) When certain nodes connect disproportionately large spatial regions, this induces undesirable load oscillations. (d) Virtual graph Voronoi centers instead of actual robot positions and dual nodes of overloaded nodes are employed to mitigate the excessive oscillations.}
	\label{fig:virtual_center}
\end{figure*}


\begin{theorem}
	\label{theo:1}	Consider the weighted graph Voronoi partitions $N_i$ being updated according to Algorithm~1 at initial time $k_0$. Suppose that Assumptions~\ref{sup:iter_bound} and \ref{sup:communication} hold. Let the parameter satisfying $B_\lambda  \geq 2\gamma \mathcal{B}$. Then, there exists a time $\hat{\kappa} > k_0$ such that for all time $k > \hat{\kappa}$, every robot $i\in V$ will achieve exploration load balance, i.e., 
	\begin{equation*}
		\max_{i \in V} \Bigg\lvert \lambda\big(\bfg_i,N_i(k)\big) - \frac{1}{\lvert V \lvert} \sum_{p \in V} \lambda\big(\bfg_p,N_p(k)\big) \Bigg\rvert
		\leq D_\lambda(k) \leq B_\lambda .
	\end{equation*}
\end{theorem}

\begin{proof}
	According to Lemma~\ref{fix_dec_D}, we can find a $h \in \mathbb{N}$ that satisfies
	\begin{equation*}
		D_\lambda(k_0+h\kappa) < D_\lambda(k_0) - h\Gamma \leq B_\lambda,
	\end{equation*}
	which $h >  \frac{D_\lambda(k_0) - B_\lambda}{\Gamma}$. Let $\hat{\kappa} = k_0 + h\kappa > k_0 + \kappa\frac{D_\lambda(k_0) - B_\lambda}{\Gamma}$ and we have
	\begin{equation*}
		\begin{aligned}
			&\max_{i \in V} \Bigg\lvert \lambda\big(\bfg_i,N_i(k)\big) - \frac{1}{\lvert V \lvert} \sum_{p \in V} \lambda\big(\bfg_p,N_p(k)\big) \Bigg\rvert \\
			&\leq \Bigg\lvert  \max_{p \in V} \lambda\big(\bfg_p,N_p(k)\big) - \min_{q \in V} \lambda\big(\bfg_q,N_q(k)\big) \Bigg\rvert \\ 
			&= D_\lambda(k) \leq D_\lambda(\hat{\kappa}) \leq B_\lambda
		\end{aligned}
	\end{equation*}
	for all time $k > \hat{\kappa}$.
\end{proof}

Theorem~\ref{theo:1} has established that the exploration loads of all robots can converge to a neighborhood of the average load with radius $B_\lambda$. We now proceed to analyze the relationship between the range of parameter $B_\lambda$ and the optimal bound achievable for global balanced exploration.

The lower bound of parameter $B_\lambda$ is related to the upper bound of the load variation per iteration under Assumption~\ref{sup:iter_bound}, i.e., $B_\lambda \geq 2\gamma\mathcal{B}$. In most cases, $\mathcal{B}$ cannot be equal to zero. As illustrated in Fig.~\ref{fig:virtual_center} (a), there may exist unnecessarily partitioned small spatial primitives (i.e., regions where a single robot suffices for complete exploration). If the number of such small spatial primitives does not match the number of robots, it necessarily implies that $\mathcal{B}$ cannot be zero.

The weighted graph Voronoi performs repartition along inter-center paths. When overloaded nodes connect disproportionately large spatial regions, this induces undesirable load oscillations (see Fig.~\ref{fig:virtual_center} (b) and (c)).  
The repartition of the overloaded node may trigger cascading partition changes across all dependent nodes, potentially leading to workload oscillations.
To address this, we employ virtual graph Voronoi centers instead of actual robot positions and dual nodes of overloaded nodes to achieve globally optimized exploration area partition (Fig.~\ref{fig:virtual_center} (d)). 

\subsection{Virtual Graph Voronoi Centers and Dual Nodes}
\label{section: VirCen}
To mitigate this issue, we introduce the dual nodes.
A node $\hat\bfv$ is \emph{overloaded} at time $k$ if
\begin{equation*}
	\left\lvert \lambda\big(\bfg_i,N_i(k)\big) - \lambda\Big(\bfg_i,N_i(k)\setminus \{\hat\bfv \} \Big) \right\lvert \geq \mathcal{L},
\end{equation*}
where $\mathcal{L}$ is exploration load threshold, which a single robot suffices for complete exploration.
Then the dual nodes are sampled in the free space of the the overloaded node's $\hat\bfv$ neighborhood $\{\mathbf{x} \in \mathcal{X}_{free}| \Vert \mathbf{x} - \hat\bfv \Vert \leq \epsilon \}$, thereby establishing new connectivity relationships. 

The introduction of dual nodes enhances the flexibility of graph space partitioning at overloaded points while ensuring that perturbations caused by single-node repartitioning remain below the exploration capability threshold $\gamma \mathcal{B} \leq \mathcal{L}$. 
Notably, dual nodes are generated only at overloaded nodes $\hat\bfv$ after weighted graph Voronoi partitioning induces perturbations, rather than requiring overload evaluation for all nodes in the topological map. This approach effectively balances node quantity and reduces computational overhead in weighted graph Voronoi partitions.

Moreover, selecting endpoints of longer topological paths as graph Voronoi centers increases the adjustable graph space (inter-center paths) within the weighted graph Voronoi partition.
Based on this characteristic, we propose a virtual center selection strategy that maximizes dispersion of centers in the topological map.
For robot $i$, let $G_{ci} = G_i\bigcup_{j \in \mN_i} G_j = (N_{ci},~E_{ci})$ denote its merged topological graph with neighbors. 
Select the node $\bfv \in N_i$ that maximizes the sum of distances to all neighboring Voronoi centers as the virtual center $\bfg_i$, i.e.,
\begin{equation}
	\label{eq:vir_center}
	\bfg_i = \mathop{\mathrm{arg}\max}\limits_{\bfp \in N_i} \sum_{j \in \mN_i} \mathcal{D}^{G_{ci}}(\bfp,\bfg_j).
\end{equation}
If the solution to \eqref{eq:vir_center} is non-unique, the node closest to the current position is selected as the virtual center.

\section{Balanced collaborative online exploration}
\label{section:online_exploration}
In Section~\ref{section:graph_Voronoi}, we have proposed the weighted graph Voronoi partition method for balanced non-convex exploration space partition within known environment.
In this section, the weighted graph Voronoi partition $N_i$ is extended to a online version and a novel hybrid topological map $G^H_i$ is developed for balanced collaborative online exploration.
Guided by the topological maps $G^H_i$ with the weighted Voronoi partition $N_i$, robots select and navigate toward the next viewpoint targets with enhanced foresight and efficiency.

\begin{figure} [t]
	\centering
	\begin{overpic}[width=7.8cm,height=7.2cm,trim=0 0 0 0, clip]{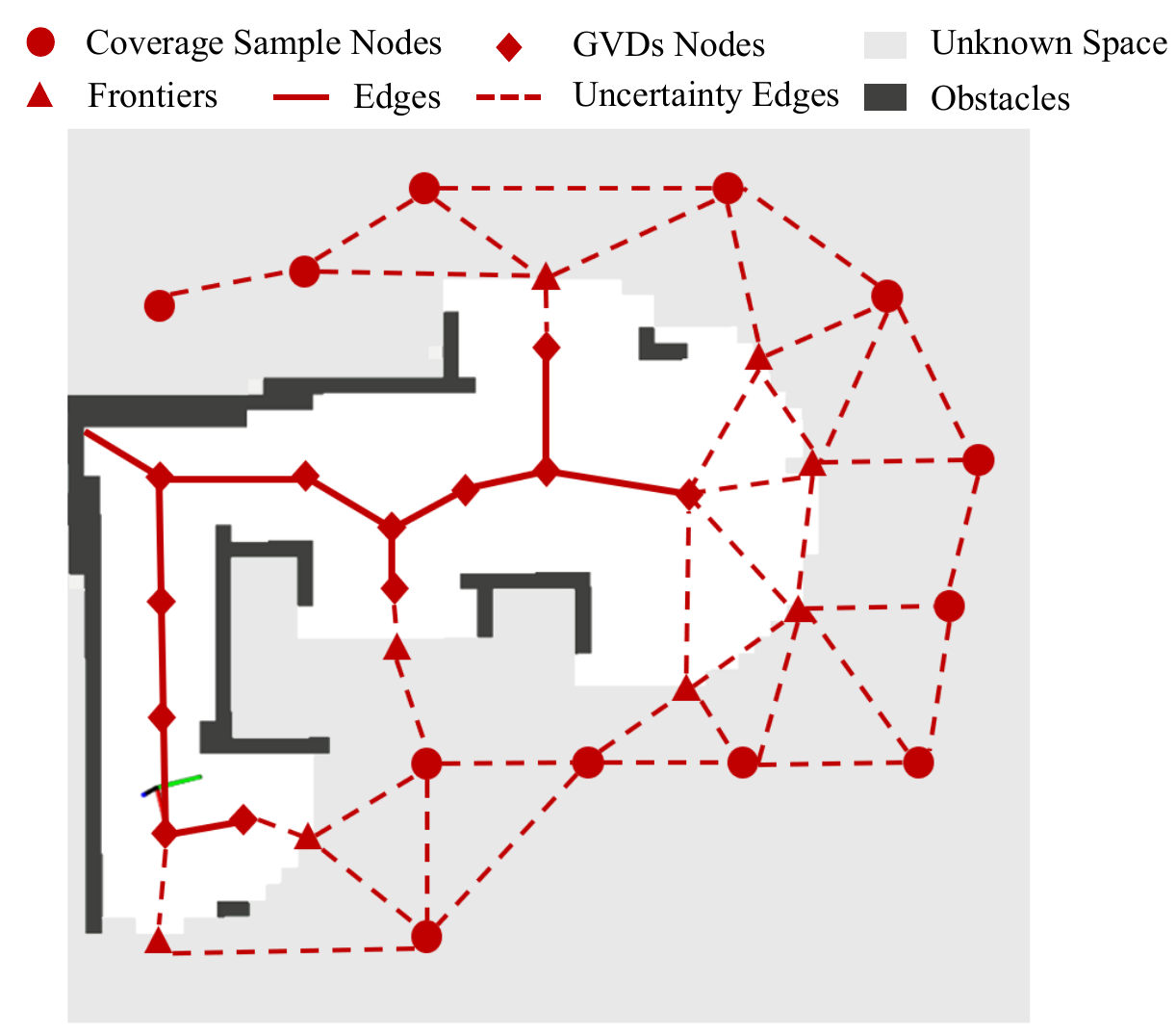}
	\end{overpic}
	\caption{Structure of the hybrid topological map.}
	\label{fig:topological_map}
\end{figure}

\subsection{System Overview}

The entire balanced online collaborative exploration planner is summarized in Fig.~\ref{fig:system_overview}.
During online exploration, robot $i$ constructs a local hybrid topological map $G^H_i$ using onboard sensors (\ref{incremental_topo}). With a fix frequency, robots communicate and share the occupancy $\mathcal{M}_i$ and topological $G^H_i$ maps with neighbors $j \in \mN_i$. 
The topological maps shared to $i$ are merged to obtain $G^H_{ci} = G^H_i\bigcup_{j \in \mN_i} G^H_j$.
Then, distributed weighted graph Voronoi partitions $N_i$ (Algorithm~\ref{al:iter_graphVor_1}) are operated until the exploration workloads are converged to balance (\ref{section:online_VC}). 
Finally, robot $i$ visits the allocated targets by hierarchical motion planner (\ref{section: motion_planner}).

\subsection{Incremental Hybrid Topological Map Construction}
\label{incremental_topo}
Each robot $i$ maintains a \emph{hybrid topological map} $G^H_i$, which is designed to  simultaneously represent exploration completeness and environmental connectivity. 
The hybrid topological map is defined as $G^H_i=(\{N^{GV}, N^{F}, N^{C}\},\{E,E^U\})$, which is illustrated in Fig.~\ref{fig:topological_map}.
\textit{Generalized Voronoi nodes} (GV nodes) $N^{GV}$ construct the environmental connectivity via generalized Voronoi diagram \eqref{eq:GVD}. 
\textit{Frontier nodes} $N^{F}$ guarantee the exploration completeness, 
and \textit{coverage sample nodes} $N^{C}$ guide foresighted coverage path planning. 
Meanwhile, there are two distinct edge types: deterministic edges $E$ connecting two \textit{GV nodes} (uniquely determined by obstacle distributions) and uncertain edges $E^U$ involving other nodes.

\begin{figure} [t]
	\centering
	\begin{overpic}[width=9cm,height=5cm,trim=0.2cm 0 0 0, clip]{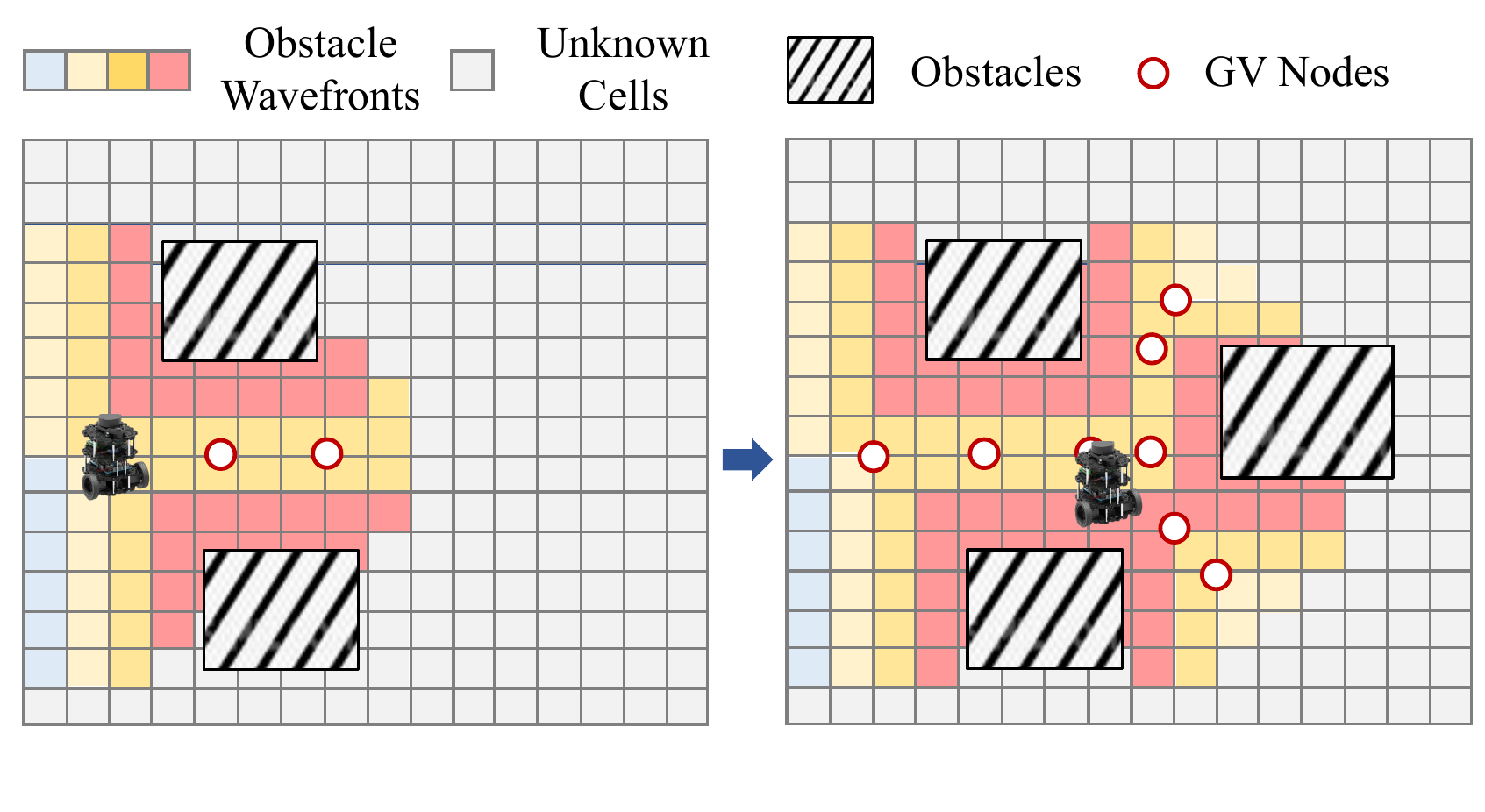}
		\put(23,1){\color[rgb]{0,0,0}{(a)}}
		\put(73,1){\color[rgb]{0,0,0}{(b)}}  
	\end{overpic}
	\caption{(a) The dynamic Brushfire algorithm for incremental generation of GV nodes. (b) Upon detecting new obstacles, the distance map of surrounding known cells is incrementally updated.}
	\label{fig:GVnodes}
\end{figure}

All the nodes in $G^H_i$ are updated incrementally during online exploration.
GV nodes $N^{GV}$ are updated by the dynamic Brushfire algorithm \cite{2009_Nidhi_ras}.
A dynamic obstacle set is maintained and all obstacles in the set propagate wavefronts until  wavefront collision or reach maximum propagation range $\hat{D}$ in already known area, which is illustrated in Fig.~\ref{fig:GVnodes}. The wavefront collision points are recorded as $N^{GV}$.
Frontier nodes $N^F$ represent the candidate viewpoints as the boundaries between explored and unexplored regions. The incremental update of frontiers follows \cite{zhou_fuel_2021}. A frontier information structure (FIS) comprises the cells within a frontier cluster, its cluster center, the viewpoints covering the cluster, and an axis-aligned bounding box. When the local map undergoes updates, the system evaluates whether any frontier clusters are affected, subsequently removing the FISs of impacted clusters while simultaneously detecting new frontiers and generating their corresponding FISs.
The coverage nodes $N^C$ are obtained by sampling the unknown regions within the sliding window of the robot, where the sample space is $\{\mathbf{x} \in \mathcal{X}_{free}| \Vert \mathbf{x} - \mathbf{r}_i \Vert \leq d_c \}$. The introduction of coverage nodes enables the exploration planning and target allocation to consider more long-term global coverage.

Differing from existing methods that construct environmental connectivity topology through A* search \cite{FALCON}, or RRT sampling \cite{vutetakis_active_2024}, the proposed hybrid topological map uses GV nodes. 
This formulation offers superior stability, as nodes are uniquely determined by obstacle configurations, significantly reducing computational overhead during distributed topological map fusions.
Moreover, distinct from the topology in Section~\ref{section:graph_Voronoi} that only employs GV nodes for known environment partition, the frontier nodes $N^F$ and coverage sample nodes $N^C$ in hybrid topological map significantly enhance long-term global exploration completeness and efficiency during online exploration.

\subsection{Online Weighted Graph Voronoi Partitions}
\label{section:online_VC}
In Section~\ref{section:graph_Voronoi}, we have proposed the weighted graph Voronoi partitions and proved in Theorem~\ref{theo:1} that Algorithm~\ref{al:iter_graphVor_1} can converge the exploration workload to equilibrium. Here we extend the Algorithm~\ref{al:iter_graphVor_1} to a online version.

The distributed topological map $G^H_{ci}$ merging and exploration workload metrics are refined through the definition of uncertain edges $E^U$. 
The merging of topological maps first computes the geometric transformation between the maps. Then, it completely preserves the deterministic edges in $E$ connecting GV nodes. 
For nodes with uncertain edges in $E^U$, i.e., nodes in $N^F$ and $N^C$,
their density is regulated based on Euclidean distances.
The nodes separated by less than a threshold distance are merged into a single node, and the connectivity of merged nodes is updated according to shared occupancy information.
Moreover, the graph-based exploration load metric \eqref{eq:load_metric} for robot $i$ is refined to the online version $\lambda^\mathcal{O}(\bfg_i,N_i)$:
\begin{equation}
	\label{eq:online_load_metric}
	\begin{aligned}
		&\mathcal{E}(\bfg_i,N_i) = \bigcup_{\bfv \in N_i} \mathcal{P}(\bfg_i, \bfv), \\
		&\lambda^\mathcal{O}(\bfg_i,N_i) = \sum_{e \in \mathcal{E}(\bfg_i,N_i)} \delta(e) \mD^{G^H_{ci}}(e), \\
		&\delta(e) =  \left\{
		\begin{array}{ll}
			1, & \text{if}~e \in E^U, \\
			0, & \text{otherwise},
		\end{array}
		\right. 
	\end{aligned}
\end{equation}
where $\delta(e)$ effectively quantifies whether the endpoints of uncertain edges $E^U$ requires further exploration. All the endpoints of uncertain edges in the weighted Voronoi partitions $N_i$ construct the target points $T_i \subseteq N_i$ for robot $i$ to be explored.
Note that the online exploration load metric \eqref{eq:online_load_metric} keeps the strictly decreasing property \eqref{eq:leaf_decrease}.

\begin{lemma}
	\label{fix_dec_D_online}
	Consider the weighted graph Voronoi partitions $N_i$ being updated according to Algorithm~\ref{al:iter_graphVor_1} using the online exploration load metric \eqref{eq:online_load_metric} instead of \eqref{eq:load_metric}, both Lemma~\ref{fix_dec_D} and Theorem~\ref{theo:1} remain valid.
\end{lemma}

\begin{proof}
	The proof of Lemma~2 is shown in Appendix B.
\end{proof}

\subsection{Hierarchical Motion Planner}
\label{section: motion_planner}
The balanced Voronoi partitions $N_i$ provides target points $T_i$ to be explored.
However, to enhance exploration efficiency by reducing redundant traversals and repeated explorations, it remains necessary to plan the visitation sequence for all target points $T_i$. 
Inspired by the recent works \cite{RACER}, \cite{FALCON}, we formulate a variation of restricted ATSP with coordination precedence search. 
Note that ATSP is NP-hard and it is unacceptable to compute ATSP for all nodes online. To improve computational efficiency, we design a collaborative priority metric $\pi_i(\bfv)$ to rank all target points $T_i$. 
The metric trades off between the information gain $\mathcal{I}$ and travel cost $\mathcal{C}$ similar to the information path planning methods \cite{ROAM}, \cite{2024_Jakkala_icra} 
, while incorporating a novel self-priority term $\mathcal{S}$ to enhance collaborative efficiency. The composite metric is formulated as
\begin{equation}
	\label{eq:priority_metric}
	\pi_i(\bfv) = \mathcal{I}(\bfv, \mathcal{M}_i) - \beta_\mathcal{C} \cdot \mathcal{C}(\bfv, \mathbf{r}_i, \mathcal{M}_i, G^H_{ci}) + \beta_\mathcal{S} \cdot \mathcal{S}(\bfv, G^H_{ci}),
\end{equation}
where the parameter $\beta_\mathcal{C} \in [0, 1)$ trades off between the information gain and travel cost. 
The parameter $\beta_\mathcal{S} \in [0, 1)$ determines the additional visitation priority of the nodes which only belongs to robot $i$ during the weighted Voronoi partition iterations.

The related gain functions are define as follow:
\begin{itemize}
	\item \textit{Information gain} $\mathcal{I}(\bfv, \mathcal{M}_i) \in [0,1]$ quantifies the potential new information a robot could acquire from sensor field-of-view (FoV) measurements at a given point $\bfv$, computed based on the surrounding unexplored region in $\mathcal{M}_i$. 
	The information gain $\mathcal{I}(\bfv, \mathcal{M}_i)$ and travel cost $\mathcal{C}(\bfv, \mathbf{r}_i, \mathcal{M}_i, G^H_{ci})$ are normalized to the $[0,1]$ interval using min-max normalization.
	\item \textit{Travel cost} $\mathcal{C}(\bfv, \mathbf{r}_i, \mathcal{M}_i, G^H_{ci}) \in [0,1]$ represents the motion distance from a robot's current position $\mathbf{r}_i$ to a target point $\bfv$. To improve search efficiency, we employ a dual-layer A* algorithm \cite{FALCON} operating on both topological $G^H_{ci}$ and occupancy $\mathcal{M}_i$ maps, with priority given to searches at the topological level. The system only resorts to grid searches when no feasible path can be identified through topological map $G^H_{ci}$. 
	\item \textit{Self-priority flag} $\mathcal{S}(\bfv, G^H_{ci}) \in \{0, 1 \}$ indicates whether a node exclusively belongs to robot $i$, i.e., lies far from the partition boundaries of all neighboring robots $j \in \mN_i$. 
	Nodes with exclusive partitions are granted higher visitation priority for two reasons: first, these nodes incur higher exploration costs for other robots, and second, prioritizing them can reduce repeated explorations after re-partitioning. The computation of self-priority $\mathcal{S}(\bfv, G_i)$ flag only requires marking nodes that maintain consistent partitions during weighted graph Voronoi partition updates.
\end{itemize}

All allocations $T_i$ are sorted according to the proposed collaborative priority metric \eqref{eq:priority_metric}, and a subset $T^P_i$ is greedily selected, which satisfies $T^P_i \subseteq T_i$ and $\lvert T^P_i \lvert = \min\{\Pi, \lvert T_i \lvert\}$. The parameter $\Pi$ specifies the number of nodes to be explored within the receding horizon window. Subsequently, an ATSP is solved for the subset $T^P_i$, where the cost matrix is also computed using the aforementioned dual-layer A* algorithm operating on both topological $G^H_{ci}$ and occupancy $\mathcal{M}_i$ maps to accelerate the search process. 

Furthermore, the travel distance $d^\mathcal{T}_i$ estimated by ATSP for $T^P_i$ is incorporated as feedback into the weighted graph Voronoi partition process. 
The ATSP-estimated travel distances $d^\mathcal{T}_i$ feedback helps maintain balanced actual movement distances among robots by considering the visitation sequence for target points. 
The enhanced exploration workload metric incorporating this feedback $\lambda^\mathcal{F}(\bfg_i,N_i)$ can be expressed as:
\begin{equation}
	\label{eq:feedback_load_metric}
	\lambda^\mathcal{F}(\bfg_i,N_i) = \lambda^\mathcal{O}(\bfg_i,N_i) + \gamma_d \cdot d^\mathcal{T}_i,
\end{equation}
where $\gamma_d > 0$ is a constant parameter. 
The metric $\lambda^\mathcal{O}(\bfg_i,N_i)$ represents the graph-space distance within a specified spatial range after balanced partition, while $d^\mathcal{T}_i$ denotes the ATSP travel distance considering visitation sequences within the receding horizon window. The parameter $\gamma_d$ normalizes the spatial and temporal window scales between $\lambda^\mathcal{O}(\bfg_i,N_i)$ and $d^\mathcal{T}_i$.
In practice, the parameter $\gamma_d$ can be selected as the ratio between the window length $d_c$ of $N^C$ sampling and receding horizon target number $\Pi$, i.e., $\gamma_d = \frac{d_c}{\Pi}$.
For the weighted graph Voronoi partition iteration, the travel distance feedback introduces only a constant bias term $\gamma_d\cdot~d^\mathcal{T}_i$ in initial weights, which will not affect the partition convergence. Finally, each robot generates a smooth safe trajectory using B-spline \cite{2019_zhou_ral}, ensuring constraints on the motion dynamics.

\section{Experiments and results}
\label{section:experiments}
This section presents the exploration planner evaluation environment designed to enable rigorous and unbiased simulation experiments for subsequent benchmark testing (see Section~\ref{section:sim_benchmark}) and ablation analysis (see Section~\ref{section:sim_ablation}). 
Moreover, the proposed algorithm is validated by real-world experiments (see Section~\ref{section:real_experiments}) with multiple TurtleBot3-Burger robots. 

\begin{figure} [t]
	\centering
	\begin{overpic}[width=9cm,height=9.cm,trim=0 0 0 0, clip]{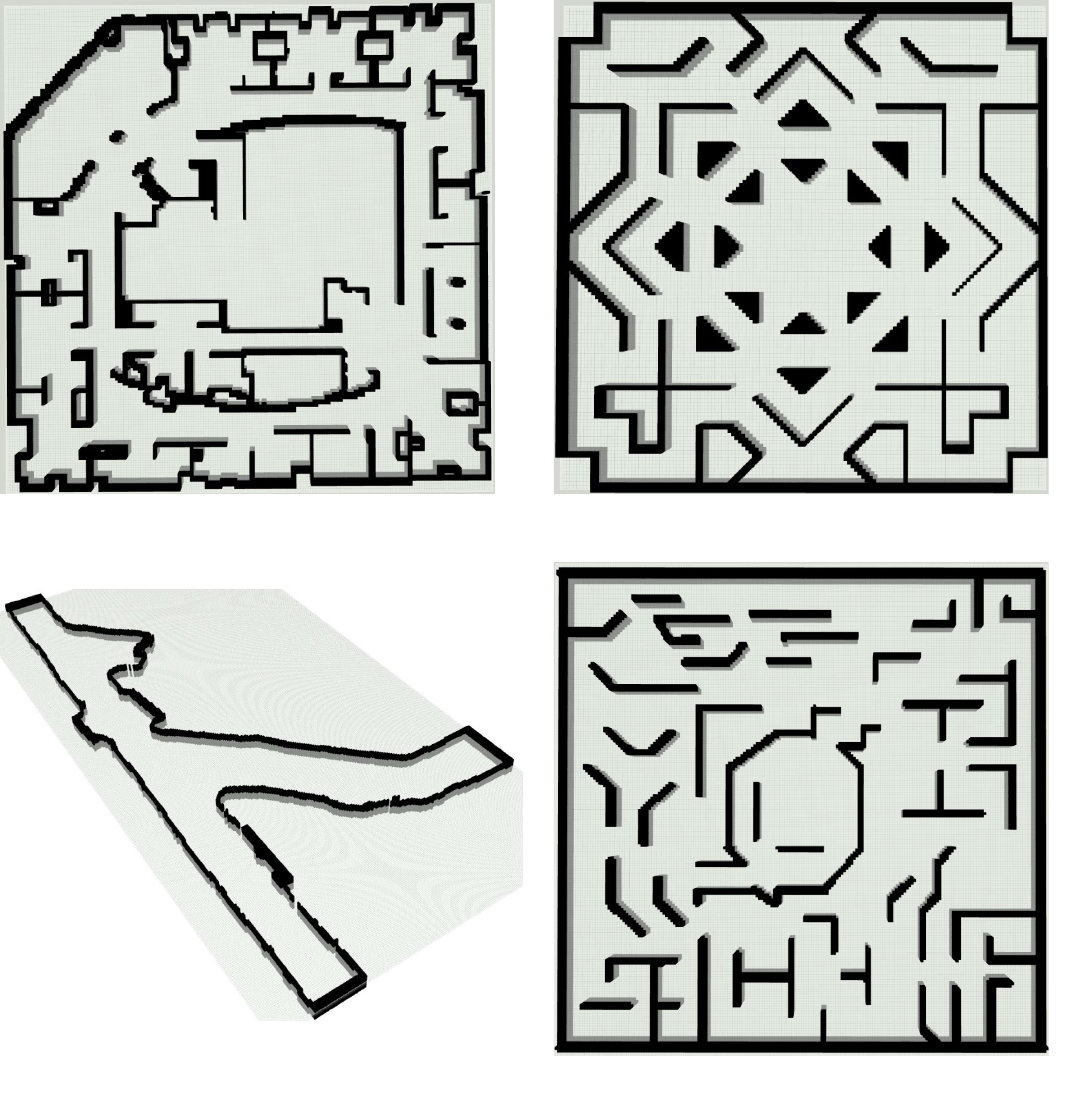}
		\put(6,51){\color[rgb]{0,0,0}{(a) Complex Office}} 
		\put(62,51){\color[rgb]{0,0,0}{(b) Octa Maze}} 
		\put(8,0){\color[rgb]{0,0,0}{(c) DARPA Tunnel}} 
		\put(62,0){\color[rgb]{0,0,0}{(d) Large Maze}} 
	\end{overpic}
	\caption{Overview of the testing scenarios used for simulation experiments. (a) Complex office ($50\text{m} \times 50\text{m}$). (b) Octa maze ($35\text{m} \times 35\text{m}$). (c) DARPA tunnel ($42\text{m} \times 20\text{m}$). (d) Large maze ($40\text{m} \times 40\text{m}$).}
	\label{fig:sim_environments}
\end{figure}
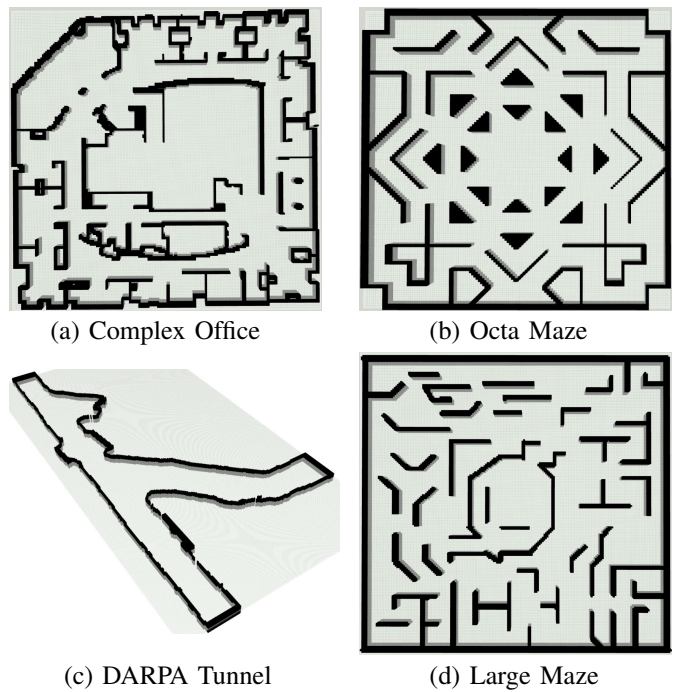

\textit{System Setup}: All simulation experiments are conducted on a computer with a AMD Ryzen 5 3600, 3.6GHz, and 48 GB RAM. 
For subsequent simulation experiments, we provide four scenarios with different levels of  complexity. To ensure fair comparison of algorithm performance, these simulation scenarios are derived from related work, including four scenarios: Complex Office, Octa Maze \cite{zhang_falcon_2025}, DARPA Tunnel \cite{zhang_falcon_2025}, and Large Maze \cite{MR-DTG}. The overviews are depicted in Fig.~\ref{fig:sim_environments}.
\begin{table}[ht]
	\centering
	\caption{Parameter Setting for Exploration Planner}
	\label{tab:para}
	\begin{tabular}{|cccc|}
		\hline
		\textbf{Type} & \textbf{Parameter} & \textbf{Notation} & \textbf{Value} \\
		\hline
		\multirow{7}{*}{Exploration Params} & Weighted iteration step & $\gamma$ & $0.5$ \\
		& Balanced threshold & $B_\lambda$ & $10$ \\
		& Travel cost weight  & $\beta_{\mathcal{C}}$ & $0.3$ \\
		& Self-priority weight  & $\beta_{\mathcal{S}}$ & $0.1$ \\
		& coverage sample window  & $d_c$ & $5.0$~m \\
		& ATSP node number  & $\Pi$ & $5$ \\
		& Actual motion feedback  & $\gamma_d$ & $1$ \\
		
		\hline
		\multirow{6}{*}{Common Params} & Max linear velocity & $v_i$ & $1.2$~m/s \\
		& Max angular velocity & $a_i$ & $1.57$~rad/s \\
		& LiDAR sensor range & - & $3.0$~m \\
		& Grid resolution & - & $0.1$~m \\
		& Safe distance threshold & - & $0.5$~m \\
		& Communication range & - & $15$~m \\
		\hline
	\end{tabular}
\end{table}

\subsection{Implementation Details}
The proposed approach employs the LKH solver \cite{HELSGAUN2000106} to address the ATSP problem described in Section~\ref{section:online_exploration}-C, while occupancy grid map updates are implemented through Gaussian process mapping \cite{2024_ding_ral}. 
All simulation experiments in Sections~\ref{section:experiments} utilize the parameter configuration specified in Table~\ref{tab:para}.
Notably, in the DARPA Tunnel environment, the LiDAR sensor range is constrained to $1.5$~meters, rendering the robots incapable of directly detecting both side walls for more challenging tests (where the minimum inter-wall distance measures $3.5$~meters).
The system executes exploration replanning at $1$~Hz and maintains inter-robot communication through $1$~Hz broadcasts within a limited range, with robots operating independently when beyond communication range and unable to exchange coordination information or map data.

\textit{Benchmark Candidates:} The benchmark comparison evaluates the proposed algorithm against a series of state-of-the-art multi-robot exploration planners, which include RACER \cite{RACER}, ROAM \cite{ROAM}, MR-DTG \cite{MR-DTG}. 
\begin{itemize}
	\item RACER \cite{RACER} employs pairwise interaction coordination to generate coverage paths for unexplored regions under capacity constraints.
	\item ROAM \cite{ROAM} implements informative path planning to optimize trajectories for simultaneous maximization of information gain and safety in collaborative semantic map estimation.
	\item MR-DTG \cite{MR-DTG} constructs a topological map structure from historical nodes and partitions the exploration area through graph Voronoi.
\end{itemize}
Note that the motion control in RACER and MR-DTG are originally designed for UAVs, with depth image point clouds as sensor input. 
For fair comparison in our simulation tests, we implement uniform local trajectory smoothing and motion control methods suitable for AGV dynamics across all benchmarked approaches, employ identical LiDAR point cloud configurations for sensor input, and adapt the information gain calculation for LiDAR-specific characteristics. 
The remaining exploration planning components are implemented using open-source codebases.

\subsection{Simulation Benchmark Results and Analysis}
\label{section:sim_benchmark}

\renewcommand{\arraystretch}{1.2}
\begin{table*}[htbp]
	\centering
	\caption{Performance Comparison}
	\label{tab:comparison}
	\begin{tabular}{|c|c|c|c|c|c|c|c|c|}
		\hline
		\multirow{2}{*}{Scene} & \multirow{2}{*}{Robot Num} & \multirow{2}{*}{Method} & \multirow{2}{*}{Exploration Time (s)}  & \multicolumn{5}{c|}{Tour Distance for All Robots (m)} \\ \cline{5-9}
		&  &  & & Avg  & Max & Min & Std & Max-Min\\ \hline
		
		\multirow{4}{*}{Complex Office} & \multirow{4}{*}{3} & 
		RACER \cite{RACER} & $176.06$ & $167.62$  & $173.35$ & $163.12$ & $4.26$ & 10.23\\ 
		& & ROAM \cite{ROAM} & $169.11$ & $161.45$   & $165.81$ & $156.78$ & $3.69$ & 9.03\\ 
		& & MR-DTG \cite{MR-DTG} & $141.74$ & $135.22$ & $138.14$ & $131.64$ & $2.69$ & 6.5\\ 
		&  & Proposed & \textbf{113.31} & \textbf{115.62} & \textbf{116.25}& 114.68 & \textbf{0.67} & \textbf{1.57} \\ \cline{2-9}
		
		\hline
		
		\multirow{4}{*}{Octa Maze} & \multirow{4}{*}{5} & 
		RACER \cite{RACER} & 99.42 & 86.59 & 99.41 & 70.13 & 13.27 & 29.28\\ 
		& & ROAM \cite{ROAM} & 90.15 & 80.04  & 88.41 & 71.47 & 5.55 & 16.94\\ 
		& & MR-DTG \cite{MR-DTG} & 86.67 & 84.54 & 85.71 & 82.18 & 1.36  & 3.53\\ 
		&  & Proposed & \textbf{72.07} & \textbf{72.09} & \textbf{72.91} & 71.33 & \textbf{0.62} & \textbf{1.58} \\ \cline{1-9}
		
		\multirow{4}{*}{DARPA Tunnel} & \multirow{4}{*}{3} & 
		RACER \cite{RACER} & 83.08 & 78.87 & 82.77 & 73.19 & 3.08 & 9.58\\ 
		& & ROAM \cite{ROAM} & 88.55 & 80.39 & 83.19  & 77.49  & 2.32 & 5.7 \\ 
		& & MR-DTG \cite{MR-DTG} & 74.86 & 75.14 & 77.36 & 73.23 & 1.71 & 4.13 \\ 
		&  & Proposed & \textbf{64.14} & \textbf{65.86} & \textbf{67.59} & 63.81 & \textbf{1.56} & \textbf{3.78} \\ \cline{1-9}
		
		\multirow{4}{*}{Large Maze} & \multirow{4}{*}{6} 
		& RACER \cite{RACER} & 123.18 & 120.97 & 136.43 & 92.95 & 18.25 & 43.48\\ 
		& & ROAM \cite{ROAM} & 109.25 & 116.72  & 122.16 & 111.17 & 3.71 & 10.99\\ 
		& & MR-DTG \cite{MR-DTG} & 105.53 & 109.01  & 112.76 & 102.59 & 3.36 & 10.17\\ 
		&  & Proposed & \textbf{93.91} & \textbf{97.93}  & \textbf{99.56} & 96.12 &\textbf{1.36} & \textbf{3.44}\\ \cline{1-9}
	\end{tabular}
\end{table*}

\begin{figure*} [ht]
	\centering
	\begin{overpic}[width=16cm,height=4.cm,trim=0 0 0 0, clip]{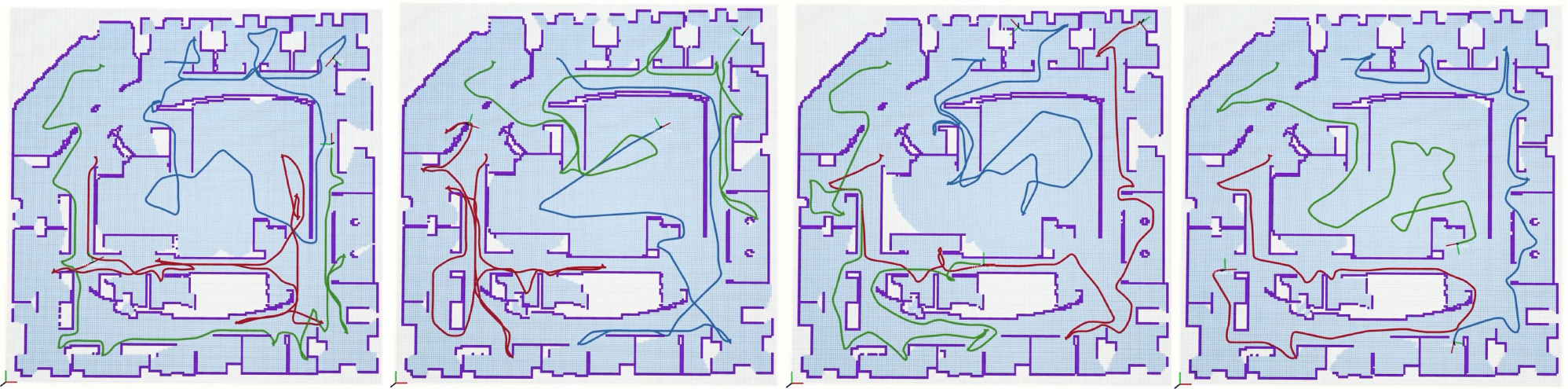}
	\end{overpic}
	\\
	\vspace{0.3cm}
	\begin{overpic}[width=16cm,height=4.cm,trim=0 0 0 0, clip]{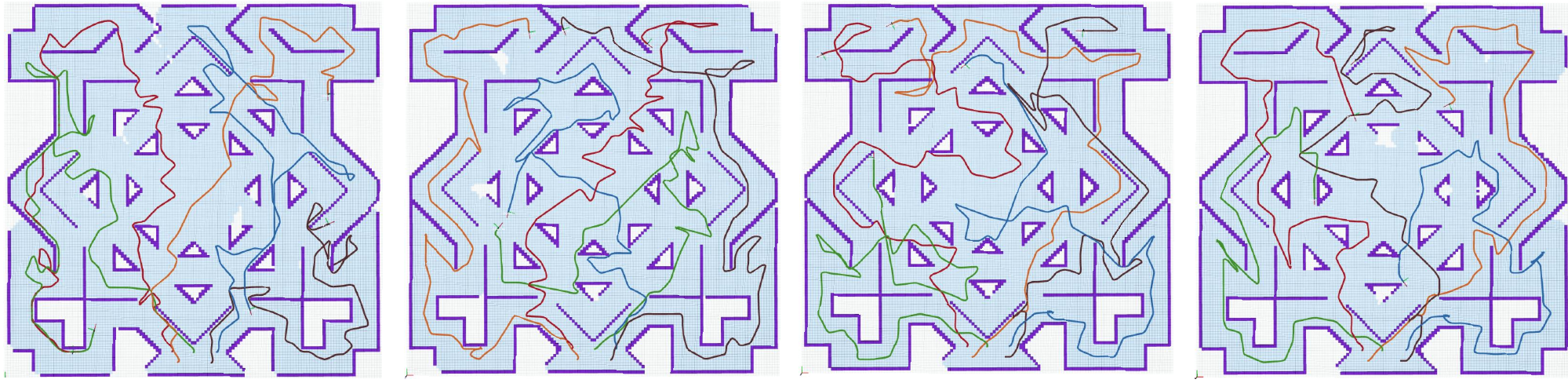}
	\end{overpic}
	\\
	\vspace{0.1cm}
	\begin{overpic}[width=16cm,height=2.5cm,trim=0 0 0 0, clip]{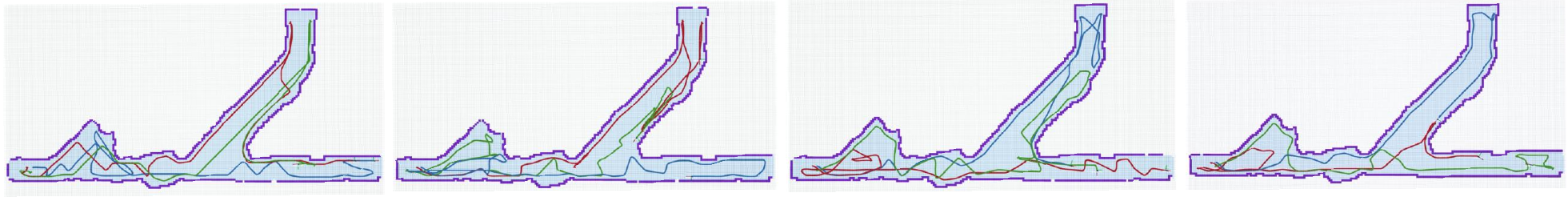}
	\end{overpic}
	\\
	\vspace{0.0cm}
	\begin{overpic}[width=16cm,height=4.cm,trim=0 0 0 0, clip]{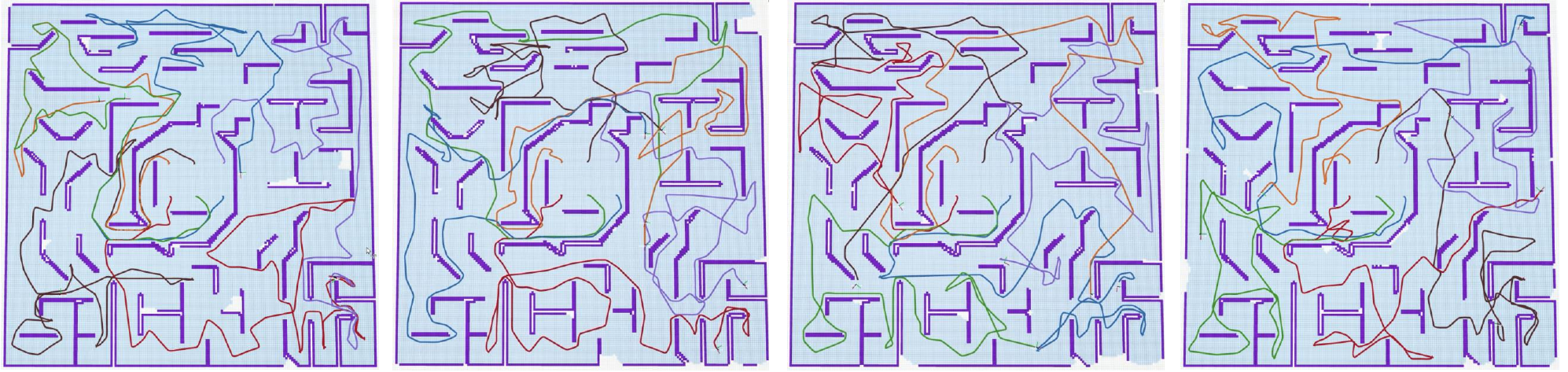}
		\put(7,-2){\color[rgb]{0,0,0}{(a) RACER}} 
		\put(31,-2){\color[rgb]{0,0,0}{(b) ROAM}} 
		\put(55,-2){\color[rgb]{0,0,0}{(c) MR-DTG}} 
		\put(82,-2){\color[rgb]{0,0,0}{(d) Proposed}} 
	\end{overpic}
	\\
	\vspace{0.25cm}
	\caption{The mapping results and the executed trajectory of all the methods. The color of the trajectory denotes different robots. Each row illustrates the simulations in one environment. From top to bottom, the scenarios are arranged sequentially as \textit{Complex Office}, \textit{Octa Maze}, \textit{DARPA Tunnel}, and \textit{Large Maze}. (a) RACER. (b) ROAM. (c) MR-DTG. (d) Proposed.}
	\label{fig:sim_results}
\end{figure*}

All the three state-of-the-art benchmarks and the proposed balanced collaborative exploration planner are comprehensively evaluated across all four scenarios. 
For each planner in each scenario, the initial states of all robots are the same. 
The statistical summary of exploration performance is presented in Table~\ref{tab:comparison}, where the exploration time refers to the duration required by all exploration planners to achieve $98\%$ environment coverage. The tour distance metrics, recorded at this exploration time, include the mean, maximum, minimum, and standard deviation of movement distances across all robots in the team.
The final mapping results and trajectories are visualized in Fig.~\ref{fig:sim_results}
and the exploration processes of environment coverage are plotted in Fig.~\ref{fig:sim_value}.
The complete exploration experiments are available in the the video of the supplementary material.
For comparative analysis of benchmark results, we focus on the efficiency and completeness of collaborative exploration, and the balance of exploration workload for multi-robot systems.

For the \textit{efficiency and completeness of collaborative exploration} criteria, each exploration planner is assessed based on the exploration time when achieve the same $98\%$ environment coverage in Table~\ref{tab:comparison}. RACER employs pairwise interaction coordination to generate coverage paths considering unknown regions under capacity constraints. 
However, RACER frequently produces back-and-forth motions reducing exploration efficiency, particularly in complex environments with dense obstacles. 
For example, this limitation manifests clearly in both the red trajectory through the \textit{Complex Office} environment and the black trajectory through the \textit{Large Maze} environment, as illustrated in Fig.~\ref{fig:sim_results}. 
This limitation stems from two fundamental issues: (1) the pairwise-update mechanism for inter-robot collaborative planning becomes unstable when neighboring relationships dynamically change during motion, frequently causing significant path replanning; 
and (2) the exploration cost metric employs Euclidean distance without accounting for actual travel distances in obstacle-dense environments, consequently inducing repetitive circumnavigation around obstacles.

ROAM implements information path planning through distributed optimization and estimation. 
Instead of partitioning the exploration space, it directly adopts a greedy strategy to select the optimal exploration frontier in the current distributed estimation. 
However, ROAM fails to ensure exploration completeness and lacks prospective planning for unknown spaces. 
Consequently, the robots frequently revisit partially explored areas, leading to progressively deteriorating exploration efficiency as illustrated in Fig.~\ref{fig:sim_value}.

MR-DTG exhibits enhanced exploration target allocation capability through Euclidean-to-graph space mapping. 
Nevertheless, its graph-space partition depends exclusively on instantaneous robot positions, which significantly degrades collaborative efficiency. 
This limitation is clearly manifested in the \textit{Complex Office} environment of Fig.~\ref{fig:sim_results}, where the green trajectory becomes conspicuously confined to corner areas by the red trajectory. 
Additionally, the topological graph's exclusive consideration of historical nodes results in insufficient global guidance for unknown region exploration.

Fig.~\ref{fig:sim_value} reveals that all benchmark methods exhibit varying degrees of the long-tail effect when pursuing $98\%$ coverage, characterized by progressively diminishing exploration efficiency. This phenomenon arises from inadequate global coverage guidance, which forces robots to repeatedly revisit partially explored regions, coupled with an  inability to maintain dynamic balancing in multi-robot collaboration within unknown environments. The system suffers from exploration load imbalance where some robots complete their exploration while others still possess substantial exploration tasks, consequently degrading overall coordination efficiency. The proposed balanced collaborative exploration planner mitigates this long-tail problem and outperforms all the benchmarks in all scenarios, achieving the shortest exploration time and tour distance. The proposed method is $11.1\%-20.05\%$ faster than the runner-up. 
The superior performance originates from the non-myopic planning for unexplored regions and guaranteed-balanced collaborative exploration.

For the \textit{balance of exploration workload} criteria, each exploration planner is evaluated based on both the standard deviation of all robots' tour distances and the difference between their maximum and minimum values, as specified in Table~\ref{tab:comparison}. 
The exploration completion time is determined by the last robot to cease movement. When the robot with the minimal travel distance stops moving, the collaborative exploration efficiency deteriorates. 
Given the similar velocities across robots, the difference between maximum and minimum tour distances approximately quantifies the additional exploration time incurred due to coordination imbalance.
Note that MR-DTG employs a strategy where robots directly navigate to neighboring robots' exploration frontiers when no local frontiers are available, resulting in comparatively larger minimum tour distances than other methods. However, this approach tends to generate redundant paths through repeated exploration, leading to suboptimal coordination efficiency.
RACER exhibits rapidly escalating exploration imbalance with increasing robot team sizes due to its pairwise collaboration mechanism, as demonstrated in both the structurally symmetric environment \textit{Octa Maze} and the obstacle-dense complex environment \textit{Large Maze}.
ROAM lacks explicit optimization for exploration balance in its algorithmic design.
The proposed method ensures balanced exploration task allocation through weighted graph Voronoi partition, achieving the smallest standard deviation and minimum-maximum difference across all scenarios.

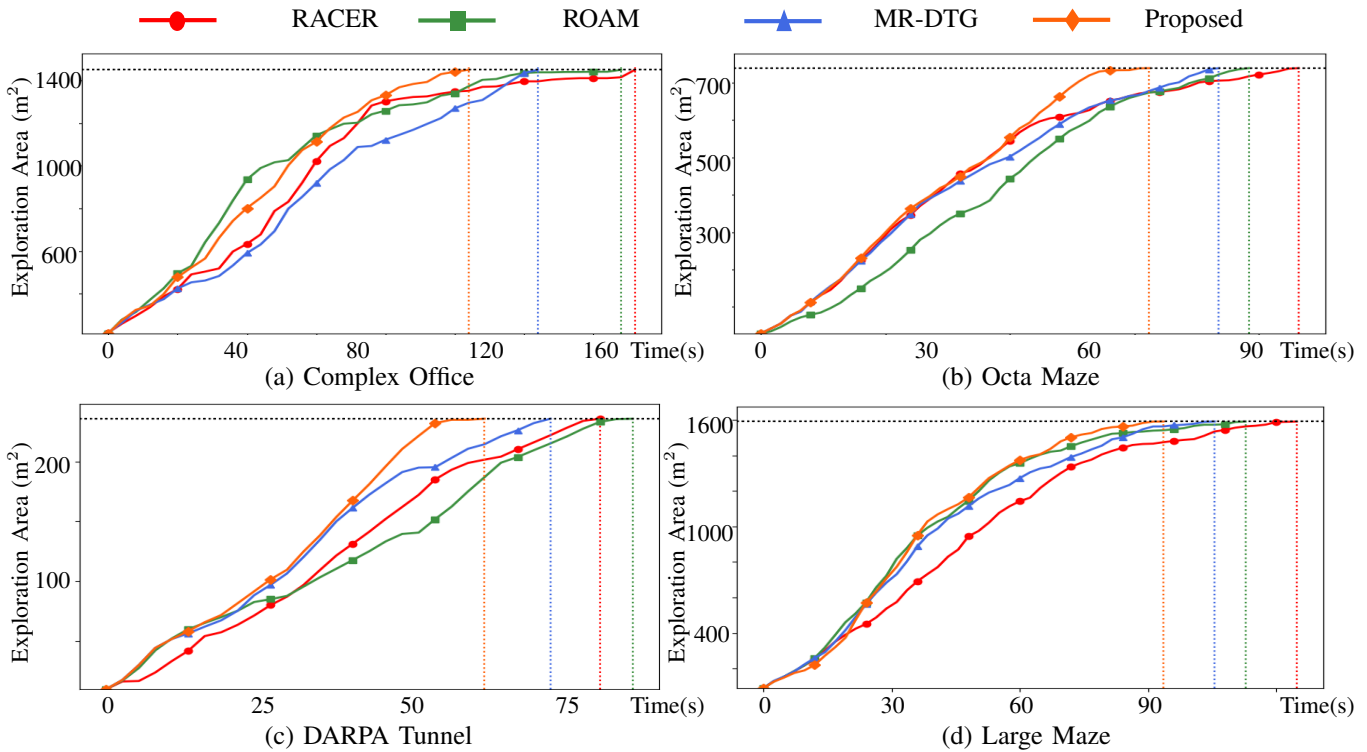
\begin{figure*} [t]
	\centering
	\begin{overpic}[width=18cm,height=10.cm,trim=0 0 0 0, clip]{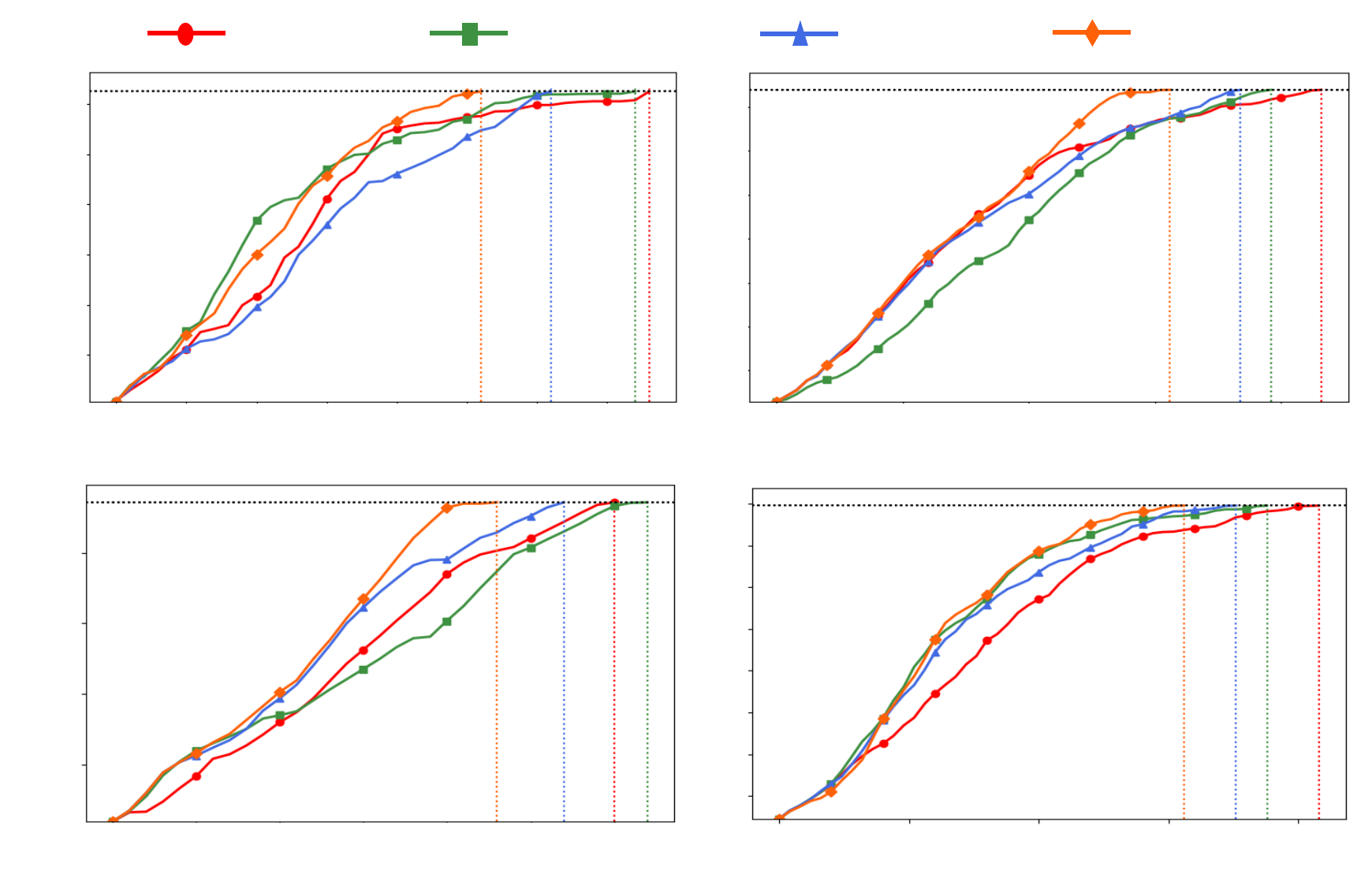}
		\put(20,26.5){\color[rgb]{0,0,0}{(a) Complex Office}}
		\put(70,26.5){\color[rgb]{0,0,0}{(b) Octa Maze}}
		\put(20,0){\color[rgb]{0,0,0}{(c) DARPA Tunnel}}
		\put(70,0){\color[rgb]{0,0,0}{(d) Large Maze}}
		\put(22,53){\color[rgb]{0,0,0}{RACER}} 
		\put(42,53){\color[rgb]{0,0,0}{ROAM}} 
		\put(65,53){\color[rgb]{0,0,0}{MR-DTG}}
		\put(85,53){\color[rgb]{0,0,0}{Proposed}}
		
		\put(1,6){\rotatebox{90}{\small Exploration Area ($\text{m}^2$)}}
		\put(1,33){\rotatebox{90}{\small Exploration Area ($\text{m}^2$)}}
		\put(49.5,6){\rotatebox{90}{\small Exploration Area ($\text{m}^2$)}}
		\put(50,33){\rotatebox{90}{\small Exploration Area ($\text{m}^2$)}}
		
		\put(3,48.5){\color[rgb]{0,0,0}{\small 1400}}
		\put(3,42){\color[rgb]{0,0,0}{\small 1000}}
		\put(3.5,35.5){\color[rgb]{0,0,0}{\small 600}}
		\put(8,28.5){\color[rgb]{0,0,0}{\small 0}}
		\put(17,28.5){\color[rgb]{0,0,0}{\small 40}}
		\put(26,28.5){\color[rgb]{0,0,0}{\small 80}}
		\put(35,28.5){\color[rgb]{0,0,0}{\small 120}}
		\put(43.5,28.5){\color[rgb]{0,0,0}{\small 160}}
		\put(47,28.5){\color[rgb]{0,0,0}{\small Time(s)}}
		
		\put(8,2.3){\color[rgb]{0,0,0}{\small 0}}
		\put(19,2.3){\color[rgb]{0,0,0}{\small 25}}
		\put(30,2.3){\color[rgb]{0,0,0}{\small 50}}
		\put(41.5,2.3){\color[rgb]{0,0,0}{\small 75}}
		\put(47,2.3){\color[rgb]{0,0,0}{\small Time(s)}}
		\put(3,20){\color[rgb]{0,0,0}{\small 200}}
		\put(3,11.5){\color[rgb]{0,0,0}{\small 100}}
		
		\put(56.2,28.5){\color[rgb]{0,0,0}{\small 0}}
		\put(68,28.5){\color[rgb]{0,0,0}{\small 30}}
		\put(80,28.5){\color[rgb]{0,0,0}{\small 60}}
		\put(92,28.5){\color[rgb]{0,0,0}{\small 90}}
		\put(95,28.5){\color[rgb]{0,0,0}{\small Time(s)}}
		\put(51.8,48){\color[rgb]{0,0,0}{\small 700}}
		\put(51.8,42.3){\color[rgb]{0,0,0}{\small 500}}
		\put(51.8,37){\color[rgb]{0,0,0}{\small 300}}
		
		\put(56.5,2.3){\color[rgb]{0,0,0}{\small 0}}
		\put(65.5,2.3){\color[rgb]{0,0,0}{\small 30}}
		\put(75,2.3){\color[rgb]{0,0,0}{\small 60}}
		\put(84.5,2.3){\color[rgb]{0,0,0}{\small 90}}
		\put(95,2.3){\color[rgb]{0,0,0}{\small Time(s)}}
		\put(51.1,23){\color[rgb]{0,0,0}{\small 1600}}
		\put(51.1,15){\color[rgb]{0,0,0}{\small 1000}}
		\put(51.7,7.5){\color[rgb]{0,0,0}{\small 400}}
	\end{overpic}
	\caption{Exploration progress of all the three state-of-the-art benchmarks and the proposed exploration planner in all the four testing scenarios. (a) Complex Office. (b) Octa Maze. (c) DARPA Tunnel. (d) Large Maze.}
	\label{fig:sim_value}
\end{figure*}

\subsection{Ablation Analysis}
\label{section:sim_ablation}
Through comprehensive ablation studies in simulation, we validate the proposed method by quantitatively evaluating the complete exploration planner against baseline variants, thereby demonstrating the effectiveness of each individual module.

According to the proposed modules, we design the following baseline variants for comparison:
\begin{enumerate}
	\item \textit{NoWeight}: This variant implements standard graph Voronoi partition without the weight bias term $w_{ij}$ in Eq.~\eqref{WGV}.
	\item \textit{PosVor}: This variant employs our weighted graph Voronoi approach but uses actual robot positions $\mathbf{r}_i$ rather than the virtual centers from Eq.~\eqref{eq:vir_center}.
	\item \textit{NoFB}: This variant modifies the weight iteration metric by excluding the travel distance feedback component specified in Eq.~\eqref{eq:feedback_load_metric}.
\end{enumerate}
We evaluate the proposed method against baseline variants in the \textit{Large Maze} environment using identical configurations from Table~\ref{tab:para}, with comparative results presented in Table~\ref{tab:Ablation}. The complete system demonstrates superior performance, achieving both the shortest exploration duration and most balanced workload distribution, thereby validating the individual contributions of each module to overall exploration efficiency. These results quantitatively confirm that the integrated approach yields optimal performance compared to partial implementations.

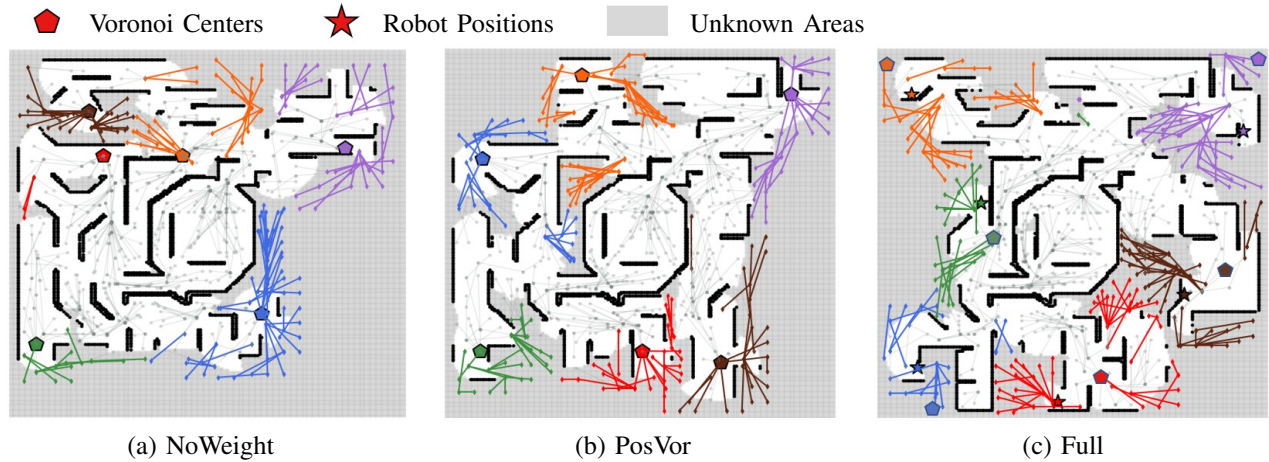
\begin{figure*} [t]
	\centering
	\begin{overpic}[width=17cm,height=6.cm,trim=0 0 0 0, clip]{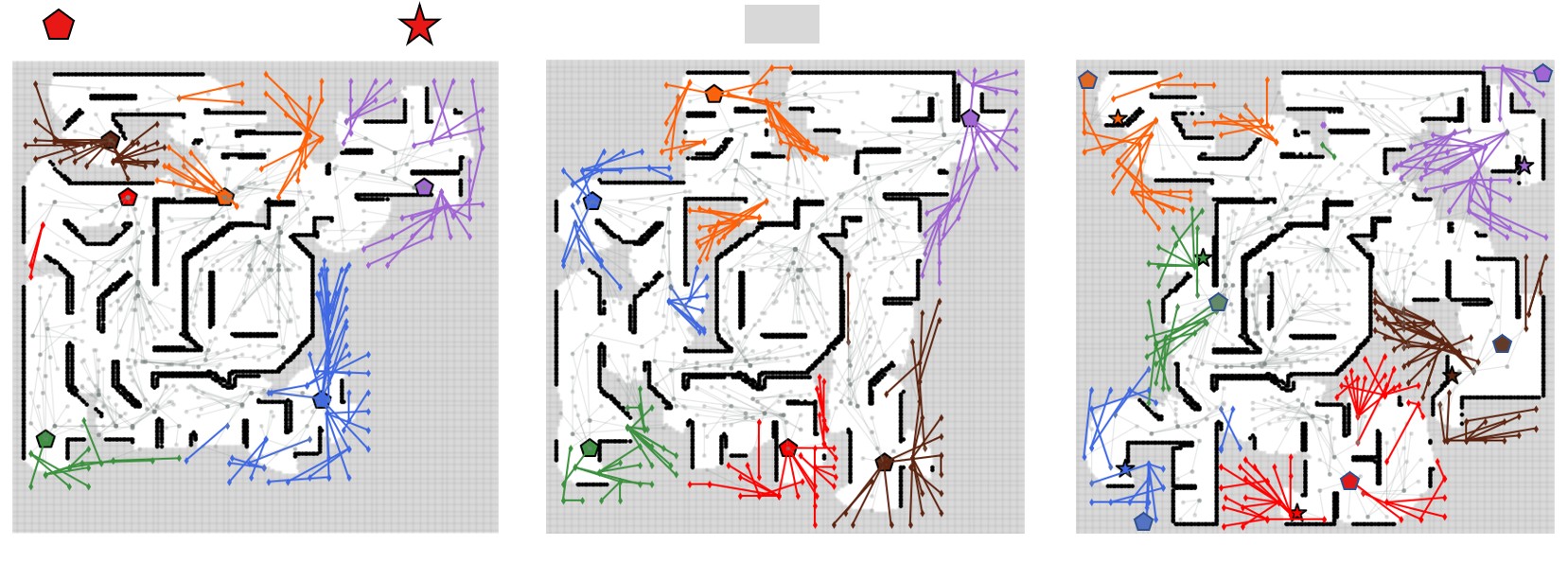}
		\put(10,0){\color[rgb]{0,0,0}{(a) NoWeight}}
		\put(45,0){\color[rgb]{0,0,0}{(b) PosVor}}
		\put(80,0){\color[rgb]{0,0,0}{(c) Full}}
		\put(7,33){\color[rgb]{0,0,0}{Voronoi Centers}}
		\put(30,33){\color[rgb]{0,0,0}{Robot Positions}}
		\put(54,33){\color[rgb]{0,0,0}{Unknown Areas}}
	\end{overpic}
	\caption{Graph Voronoi partitions at time $t=50$~s in the \textit{Large Maze} environment. Topological elements are color-coded to indicate their current partition, while gray elements represent explored areas maintained for graph connectivity. Pentagons denote current Voronoi centers $\bfg_i$, and stars mark robot positions $\mathbf{r}_i$. Both (a) NoWeight and (b) PosVor directly utilize robot positions as Voronoi centers, $\bfg_i = \mathbf{r}_i$, whereas (c) Full employs virtual centers.}
	\label{fig:sim_topo}
\end{figure*}

Fig.~\ref{fig:sim_topo} illustrates the comparative performance of graph Voronoi partition between the \textit{NoWeight}, \textit{PosVor} variants and the proposed method.
Without the weight bias term, the \textit{NoWeight} variant only determines the graph space partition according to the current positions, causing exploration imbalance. The red robot has nearly completed its assigned exploration, the blue robot retains substantial unexplored areas. 
The \textit{PosVor} variant uses actual robot positions as Voronoi centers, $\bfg_i = \mathbf{r}_i$. When robots approach each other, the weight iteration performance degradation, leading to localized imbalance. 
In contrast, our proposed virtual center formulation Eq.~\eqref{eq:vir_center} maintains optimal balance throughout exploration by maximizing inter-center distances, effectively decoupling partition quality from robot movement dynamics. This approach consistently achieves balanced exploration area partition regardless of robot positional changes.

\begin{figure} [t]
	\centering
	\begin{overpic}[width=8.8cm,height=5cm,trim=0 0 0 0, clip]{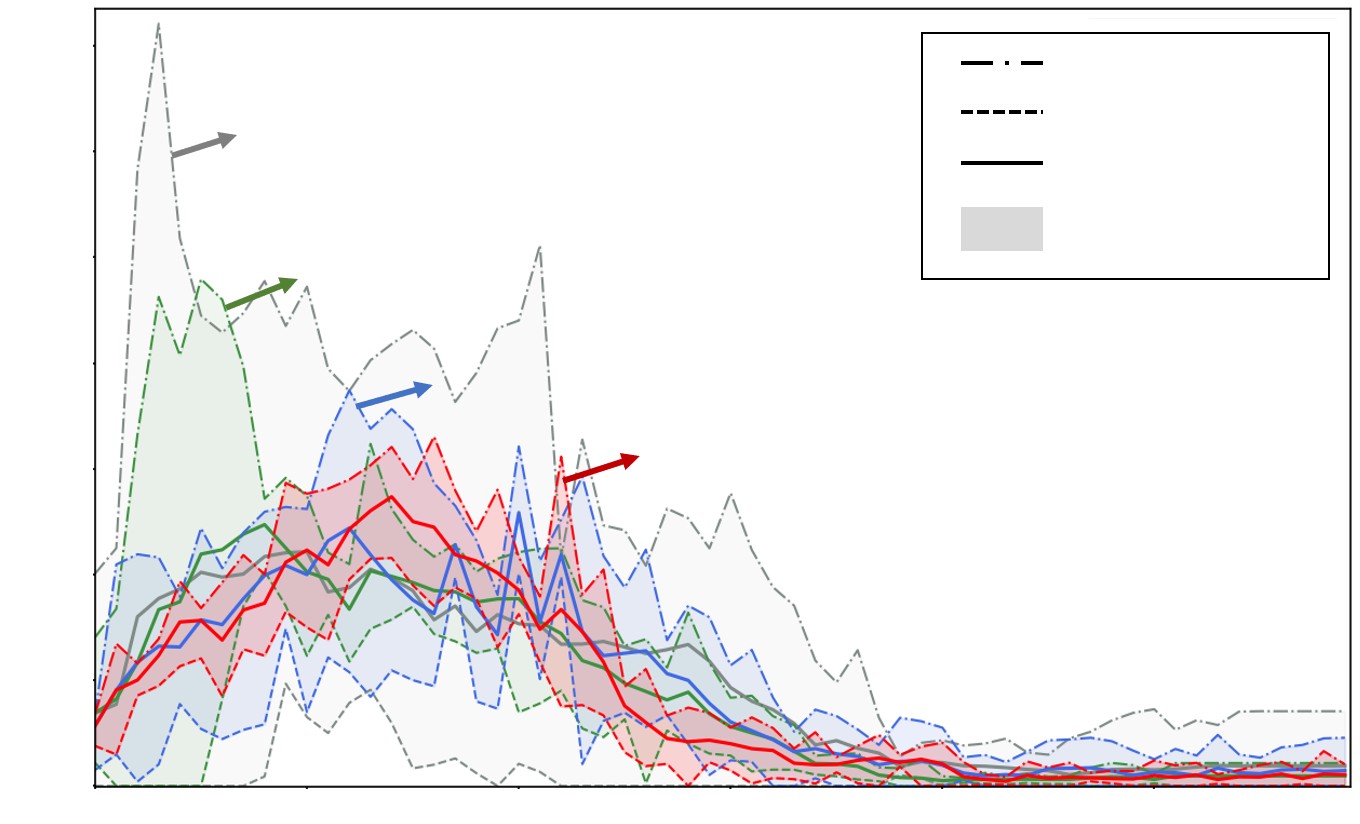}
		\put(18,45){\color[rgb]{0,0,0}{NoWeight}}
		\put(22,36){\color[rgb]{0,0,0}{PosVor}}
		\put(33,29){\color[rgb]{0,0,0}{NoFB}}
		\put(48,24){\color[rgb]{0,0,0}{Full}}
		
		\put(79,51){\color[rgb]{0,0,0}{\small Max Load}}
		\put(79,47){\color[rgb]{0,0,0}{\small Min Load}}
		\put(79,43){\color[rgb]{0,0,0}{\small Mean Load}}
		\put(79,39){\color[rgb]{0,0,0}{\small Max-Min}}
		
		\put(-1,15){\rotatebox{90}{\small Exploration Loads $\lambda^\mathcal{O}$}}
		\put(2.5,12){\color[rgb]{0,0,0}{\small 20}}
		\put(2.5,24){\color[rgb]{0,0,0}{\small 40}}
		\put(2.5,36){\color[rgb]{0,0,0}{\small 60}}
		\put(2.5,48){\color[rgb]{0,0,0}{\small 80}}
		
		\put(6,-1){\color[rgb]{0,0,0}{\small 0}}
		\put(29,-1){\color[rgb]{0,0,0}{\small 30}}
		\put(54,-1){\color[rgb]{0,0,0}{\small 60}}
		\put(79,-1){\color[rgb]{0,0,0}{\small 90}}
		\put(88,-1){\color[rgb]{0,0,0}{\small Time(s)}}
	\end{overpic}
	\caption{Exploration workloads of all variants in the \textit{Large Maze} environment. The complete proposed framework achieves achieves optimal balance throughout the exploration process.}
	\label{fig:sim_maxmin}
\end{figure}
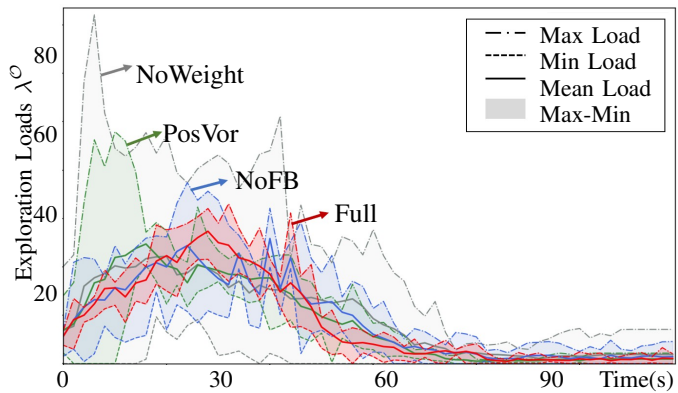

Comparative analysis of dynamic exploration workload in the \textit{Large Maze} environment is shown in Fig.~\ref{fig:sim_maxmin}. The \textit{NoWeight} variant fails to adjust exploration workload balance due to the absence of weighting terms. Both \textit{NoWeight} and \textit{PosVor} variants utilize actual robot positions as Voronoi centers, exhibiting significant partition performance degradation when initial robot positions are in close proximity. The \textit{NoFB} variant demonstrates substantially larger min-max range in exploration workload compared to the complete method \textit{Full} after removing the travel distance feedback module, indicating that incorporating online travel distance feedback effectively reduces the metric discrepancy between graph-space measurements and actual path distances considering visitation sequences.

\begin{table*}[htbp]
	\centering
	\caption{Ablation Analysis in \textit{Large Maze}}
	\label{tab:Ablation}
	\begin{tabular}{|c|c|c|c|c|c|c|c|c|}
		\hline
		\multirow{2}{*}{Scene} & \multirow{2}{*}{Robot Num} & \multirow{2}{*}{Method} & \multirow{2}{*}{ \parbox{1.2cm}{Exploration \\ ~~Time (s)} }  & \multicolumn{5}{c|}{Tour Distance for All Robots (m)} \\ \cline{5-9}
		& & & & Avg  & Max & Min & Std & Max-Min \\ \hline
		
		\multirow{4}{*}{Large Maze} & \multirow{4}{*}{6} 
		& NoWeight & 104.47 & 104.92  & 109.82 & 101.27 & 3.22 & 8.55 \\ 
		& & PosVor & 99.53 & 101.93 & 105.74 & 99.67 & 1.90 & 6.07 \\ 
		& & NoFB & 100.73 & 103.38 & 106.58 & 100.23 & 2.51 & 6.35 \\ 
		& &  Full & \textbf{93.91} & \textbf{97.93}  & \textbf{99.56} & 96.12 &\textbf{1.36} & \textbf{3.44} \\ \cline{1-9}
	\end{tabular}
\end{table*}

\subsection{Real-World Experiments}
\label{section:real_experiments}
\begin{figure} [t]
	\centering
	\begin{overpic}[width=8cm,height=4.cm,trim=0 0 0 0, clip]{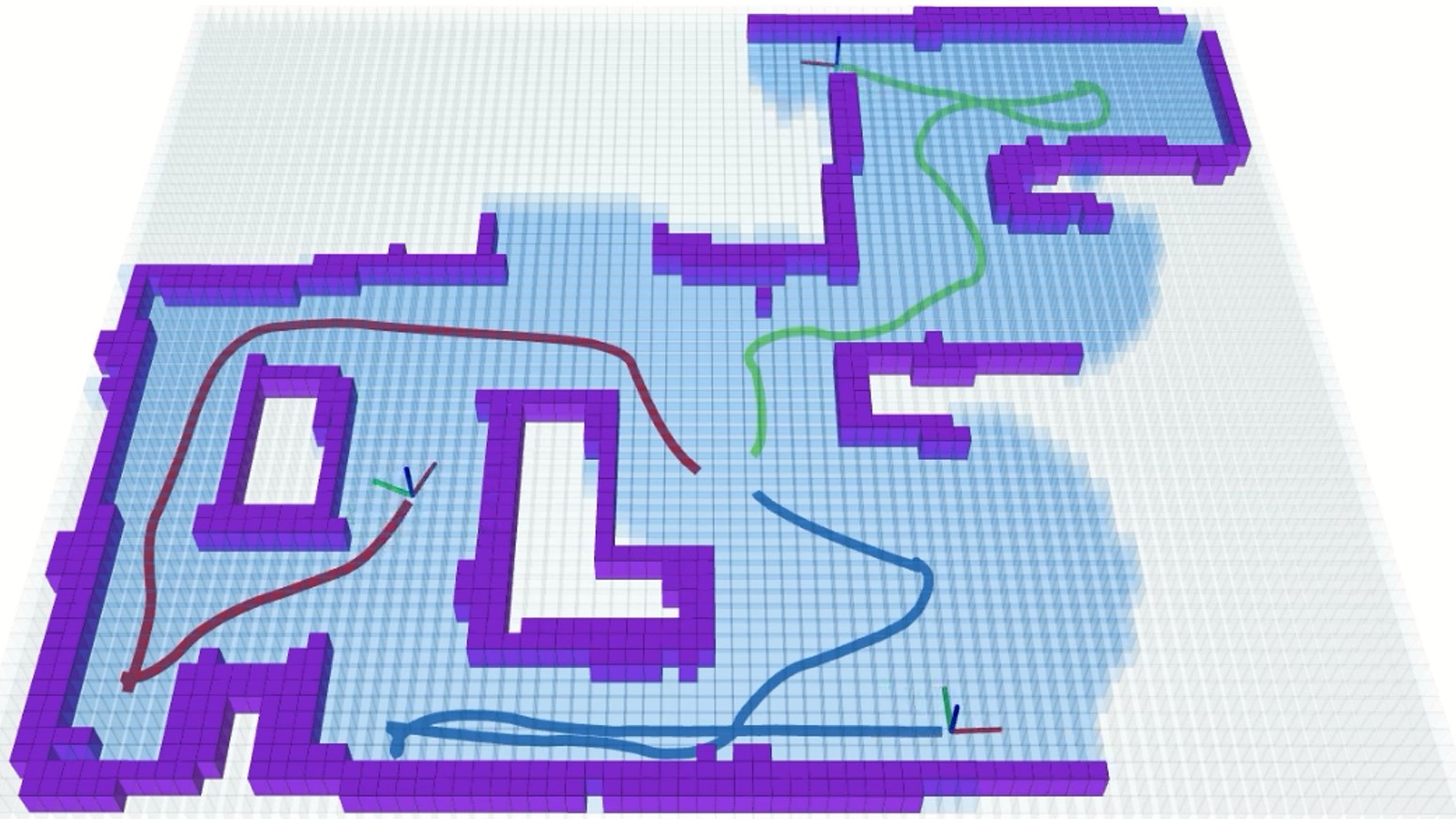}
		\put(35,-4){\color[rgb]{0,0,0}{(a) $t=15$~s}} 
	\end{overpic}
	\\
	\vspace{0.6cm}
	
	\begin{overpic}[width=8cm,height=4.cm,trim=0 0 0 0, clip]{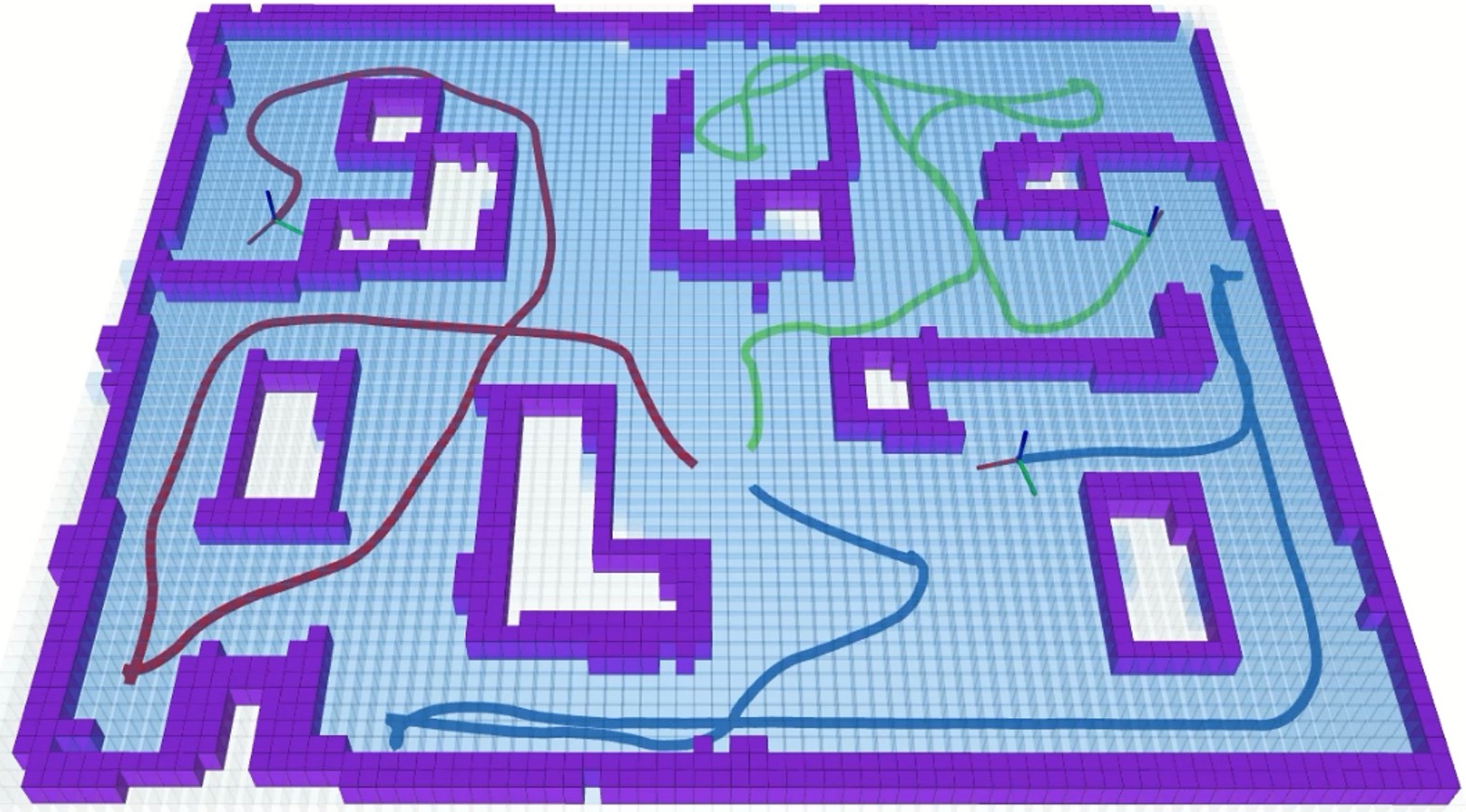}
		\put(35,-4){\color[rgb]{0,0,0}{(b) $t=28$~s}} 
	\end{overpic}
	\\
	\vspace{0.6cm}
	\begin{overpic}[width=8cm,height=4.cm,trim=0 0 0 0, clip]{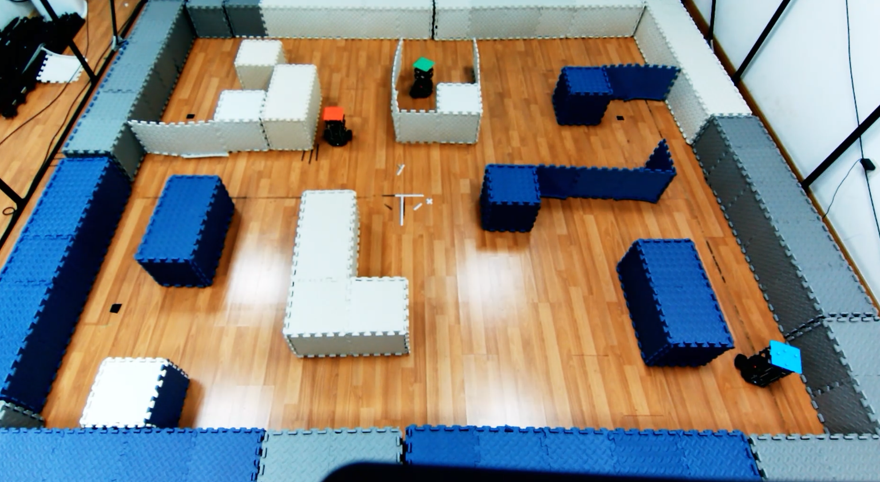}
		\put(30,-5){\color[rgb]{0,0,0}{(c) Photo of the scenario}}
	\end{overpic}
	\\
	\vspace{0.5cm}
	\caption{The mapping results and the executed trajectory in the real-world experiments.}
	\label{fig:real_world}
\end{figure}

To validate the practical feasibility of our framework for balanced collaborative exploration in real-world scenarios, we conducted field experiments using three TurtleBot3-Burger robots equipped with onboard Raspberry Pi 3B computers (Quad Core 1.2 GHz Broadcom BCM2837 CPU, 1GB RAM) operating without external infrastructure support.
The $4 \times 5~\text{m}^2$ testing environment is illustrated in Fig.~\ref{fig:real_world}(c).
The experimental setup imposed the following constraints: robot dynamics limited to maximum linear velocities of $0.4$~m/s and angular velocities of $1$~rad/s, LiDAR sensing range of $0.5$~m, safe distance threshold of $0.2$~m and inter-robot communication range of $2$~m. Other exploration parameters are the same in the Table~\ref{tab:para}. 


The experimental results demonstrate successful exploration completion in $27.6$~seconds, with three robots achieving an average tour distance of $8.44$~m (maximum: $8.53$~m, minimum: $8.35$~m, standard deviation: $0.07$~m) as shown in Fig.~\ref{fig:real_world}. These real-world validation tests conclusively verify the proposed planner's effectiveness and robustness in achieving balanced collaborative exploration fully onboard in non-convex obstacle-dense environments. Additional implementation details are available in the Supplementary Material video.

\section{Conclusion}
\label{section:conclusion}
This paper presents a distributed multi-robot exploration framework that addresses the challenge of collaborative online exploration in obstacle-dense environments. 
We introduce a novel topological mapping approach that simultaneously encodes environmental connectivity and exploration completeness, with incremental updates leveraging known spatial information. 
The system features a weighted topological graph Voronoi algorithm that guarantees bounded equitable partitions with provable convergence properties. 
Our hierarchical planning architecture combines global coverage guidance with local trajectory optimization, generating collision-free, dynamically feasible paths while minimizing travel distance. 
Extensive experimental validation demonstrates superior performance in exploration efficiency, environmental coverage completeness, and workload balance compared to state-of-the-art methods.
In future work, we will focus on further enhancing the performance of multi-robot collaborative exploration 
by incorporating semantic information and environmental prediction to improve online target assessment and allocation, 
while employing learning-based and generative approaches to optimize trajectory generation efficiency, thereby circumventing the computational constraints inherent in TSP solutions.


\section*{APPENDIX~A: Proof of the Lemma~1}
\begin{proof}
	Since the communication for all robots in $V$ is connected, we can
	define $\hat{P}(k) \subset V$ as the subset of the robots, where all robots in $\hat{P}(k)$ have the maximum score of its neighbors, i.e., $\lambda\big(\bfg_{\hat{p}},N_{\hat{p}}(k)\big) \geq \lambda\big(\bfg_j,N_j(k)\big),~\forall j \in \mN_{\hat{p}},~\forall \hat{p} \in \hat{P}(k)$. 
	Note that the robot with the maximum score is in this subset at any iteration time $k$, i.e., $\bar{p}(k) \in \hat{P}(k),~\forall k$.
	
	\textbf{1)} First, we prove by contradiction that
	for any $\hat{p} \in \hat{P}(k)$, 
	$\lambda\big(\bfg_{\hat{p}},N_{\hat{p}}(k+1)\big) \leq \lambda\big(\bfg_{\hat{p}},N_{\hat{p}}(k)\big)$ and no nodes in the neighboring graph Voronoi partition $N_j(k),~j \in \mN_{\hat{p}}$ will be added into the partition $N_{\hat{p}}(k+1)$. 
	
	Suppose that a node $\bfv \in N_j(k)$ is added into $N_{\hat{p}}(k+1)$.
	According to the definition of weighted graph Voronoi \eqref{WGV}, we have 
	\begin{equation*}
		\begin{aligned}
			&\mathcal{D}^{G_{c\hat{p}}}(\bfg_{\hat{p}},\bfv)-w_{{\hat{p}}j}(k+1) \leq \mathcal{D}^{G_{c\hat{p}}}(\bfg_j,\bfv), \\
			&\mathcal{D}^{G_{cj}}(\bfg_j,\bfv)-w_{j{\hat{p}}}(k)\leq \mathcal{D}^{G_{cj}}(\bfg_{\hat{p}},\bfv).
		\end{aligned}
	\end{equation*}
	For all $j \in \mN_{\hat{p}}$ and nodes $\bfv \in N_j(k) \cup N_{\hat{p}}(k)$, the connectivity of the topological maps $G_{cj}$ and $G_{c\hat{p}}$ remains invariant between the iteration time $k$ and $k+1$. Therefore, we have $\mathcal{D}^{G_{cj}}(\bfg_j,\bfv) = \mathcal{D}^{G_{c\hat{p}}}(\bfg_j,\bfv)$ and $\mathcal{D}^{G_{cj}}(\bfg_{\hat{p}},\bfv) = \mathcal{D}^{G_{c\hat{p}}}(\bfg_{\hat{p}},\bfv)$.
	
	Since $\lambda\big(\bfg_{\hat{p}},N_{\hat{p}}(k)\big) > \lambda\big(\bfg_j,N_j(k)\big),~\forall j\in \mN_{\hat{p}}$ holds,
	according to the weight iteration \eqref{eq:weight_update},
	we have $\Delta w_{{\hat{p}}j}(k+1) = -\gamma < 0$ and $w_{{\hat{p}}j}(k+1) = -w_{j{\hat{p}}}(k) - \gamma$. Then, we derive
	\begin{equation*}
		\begin{aligned}
			&w_{j{\hat{p}}}(k) + \gamma = -w_{{\hat{p}}j}(k+1) \leq \mathcal{D}^{G_{cj}}(\bfg_j,\bfv) - \mathcal{D}^{G_{cj}}(\bfg_{\hat{p}},\bfv) \\
			&\leq w_{j{\hat{p}}}(k),	
		\end{aligned}
	\end{equation*}
	which has led to a contradiction. No new nodes in $N_{\hat{p}}(k+1)$ will be added from $N_j(k)$, and some leaf nodes may be lost to $N_j(k+1)$. Therefore, according to \eqref{eq:leaf_decrease}, the following inequality holds:
	\begin{equation}
		\label{hatp_dec}
		\lambda\big(\bfg_{\hat{p}},N_{\hat{p}}(k+1)\big) \leq \lambda\big(\bfg_{\hat{p}},N_{\hat{p}}(k)\big).
	\end{equation}
	
	\textbf{2)} Then, we prove that before the termination condition is met for $\bar{p}$ at time $k$, the robot with the maximum score for all robots at time $k+1$ must have maximum score for its neighbor at time $k$, i.e., $\bar{p}(k+1) \in \hat{P}(k)$.
	
	For brevity, denote $p' = \bar{p}(k+1)$ here.
	According to Assumption~\ref{sup:iter_bound} and the termination condition, we have 
	\begin{equation*}
		\begin{aligned}
			&\big\lvert \lambda\big(\bfg_{j},N_j(k+1)\big) - \lambda\big(\bfg_{j},N_j(k)\big) \big\lvert \leq \gamma\mathcal{B}, \\
			& \big\lvert \lambda\big(\bfg_{p'},N_{p'}(k+1)\big) - \lambda\big(\bfg_{p'},N_{p'}(k)\big) \big\lvert \leq \gamma\mathcal{B}, \\
			&\big\lvert \lambda\big(\bfg_{p'},N_{p'}(k)\big) - \lambda\big(\bfg_{j},N_j(k)\big) \big\lvert \geq B_\lambda,~j\in \mN_{p'}(k).
		\end{aligned}
	\end{equation*}
	Note that the parameter satisfying $B_\lambda \geq 2\gamma \mathcal{B}$. Applying the absolute value inequality, we obtain
	\begin{equation*}
		\begin{aligned}
			&\big\lvert \lambda\big(\bfg_{p'},N_{p'}(k)\big) - \lambda\big(\bfg_{j},N_j(k)\big) \big\lvert \geq B_\lambda \geq 2\gamma \mathcal{B} \\
			&\geq \Bigg\lvert \lambda\big(\bfg_{p'},N_{p'}(k)\big) - \lambda\big(\bfg_{j},N_j(k)\big) - \\
			& \Big[\lambda\big(\bfg_{p'},N_{p'}(k+1)\big) -   \lambda\big(\bfg_{j},N_j(k+1) \big) \Big]\Bigg\lvert.
		\end{aligned}
	\end{equation*}
	The above inequality guarantees that 
	\begin{equation*}
		\begin{aligned}
			&\text{sign}\Big(\lambda\big(\bfg_{p'},N_{p'}(k)\big) - \lambda\big(\bfg_{j},N_j(k)\big)\Big) = \\&\text{sign}\Big(\lambda\big(\bfg_{p'},N_{p'}(k+1)\big) -   \lambda\big(\bfg_{j},N_j(k+1) \big) \Big).
		\end{aligned}
	\end{equation*}
	Since $p'$ has the maximum score such that
	$\lambda\big(\bfg_{p'},N_{p'}(k+1)\big) \geq   \lambda\big(\bfg_{j},N_j(k+1)$. We have $\lambda\big(\bfg_{p'},N_{p'}(k)\big) \geq \lambda\big(\bfg_{j},N_j(k)\big)$ holds for all $j \in \mN_{p'}(k)$, i.e., $\bar{p}(k+1) \in \hat{P}(k)$.
	
	\textbf{3)} Finally, according to $\bar{p}(k+1) \in \hat{P}(k)$ and \eqref{hatp_dec}, we can derive
	$M_\lambda(k+1) = \lambda \big(\bfg_{\bar{p}(k+1)},N_{\bar{p}(k+1)}(k+1)\big) \overset{\eqref{hatp_dec}}{\leq} \lambda\big(\bfg_{\bar{p}(k+1)},N_{\bar{p}(k+1)}(k)\big) \leq M_\lambda(k)$. Therefore, we have $M_\lambda(k+1) \leq M_\lambda(k)$ holds at any time $k$.
	
	\textbf{4)} Similarly, we have $m_\lambda(k+1) \geq m_\lambda(k)$ at any time $k$. 
	It is noteworthy that in the proofs concerning $M_\lambda$ and $m_\lambda$, the sole distinction lies in the fact that the weight for $\underline{p}$ increases monotonically, which may introduce new neighboring robots in $\mN_{\underline{p}}$ during the iterative process. 
	However, the number of robots is finite and this can be straightforwardly treated as restarting the iteration from a new initial state $k'$ without affecting $m_\lambda(k'+1) \geq m_\lambda(k')$.
	
	Therefore, we have $D_\lambda(k+1) \leq D_\lambda(k)$. Since the constant $\gamma > 0$, the equals sign doesn't always hold. Then, there exist constants $\kappa \in  \mathbb{N}^+$ and $\Gamma \in \mathbb{R}^+$ such that the following inequality
	\begin{equation*}
		D_\lambda(k+\kappa) < D_\lambda(k) - \Gamma
	\end{equation*}
	holds before the termination condition is met for all robots at time $\hat{k}$, i.e., $\lvert \Delta_\lambda(j,i,k) \lvert < B_\lambda,~\forall  k \geq \hat{k} ,~\forall i \in V,~ \forall j \in \mN_i$.
\end{proof}

\section*{APPENDIX~B: Proof of the Lemma~2}
\begin{proof}
	Note that the online exploration load metric \eqref{eq:online_load_metric} keeps the strictly decreasing property \eqref{eq:leaf_decrease}.
	Here, define 
	\begin{align*}
		\hat{P}(k) = \{\hat{p}\in V | \lambda^\mathcal{O}\big(\bfg_{\hat{p}},N_{\hat{p}}(k)\big) \geq \lambda^\mathcal{O}\big(\bfg_j,N_j(k)\big),~\forall j \in \mN_{\hat{p}}\}.    
	\end{align*}
	Following the same derivation as Step~1 in Appendix~A, the inequality
	\begin{equation*}
		\lambda^\mathcal{O}\big(\bfg_{\hat{p}},N_{\hat{p}}(k+1)\big) \leq \lambda^\mathcal{O}\big(\bfg_{\hat{p}},N_{\hat{p}}(k)\big),
	\end{equation*}
	holds for all robots $\hat{p} \in \hat{P}(k)$. 
	
	Since $\delta(e) \leq 1$, we have $\lambda^\mathcal{O}\big(\bfg_i,N_i(k+1)\big) \leq \lambda\big(\bfg_i,N_i(k+1)\big)$
	for any robot $i$. 
	During once weighted Voronoi partitions, $\delta(e)$ for all edges $e$ is invariant at any iteration $k$ and we obtain
	\begin{equation*}
		\begin{aligned}
			&\big\lvert \lambda^\mathcal{O}\big(\bfg_i,N_i(k+1)\big) - \lambda^\mathcal{O}\big(\bfg_i,N_i(k)\big) \big\lvert =  \\ 
			&\Bigg\lvert \sum_{e \in \mathcal{E}(\bfg_i,N_i(k+1))} \delta(e) \mD^{G^H_{ci}}(e) - \sum_{e \in \mathcal{E}(\bfg_i,N_i(k))} \delta(e) \mD^{G^H_{ci}}(e) \Bigg\lvert \\
			& = \Bigg\lvert \sum_{e \in \Delta\mathcal{E}(\bfg_i,N_i,k+1)} \delta(e) \mD^{G^H_{ci}}(e)  \Bigg\lvert,
		\end{aligned}
	\end{equation*}
	where $\Delta\mathcal{E}(\bfg_i,N_i,k+1) = \Big(\mathcal{E}\big(\bfg_i,N_i(k+1)\big) \cup \mathcal{E}\big(\bfg_i,N_i(k)\big)\Big) \setminus \Big(\mathcal{E}\big(\bfg_i,N_i(k+1)\big) \cap \mathcal{E}\big(\bfg_i,N_i(k)\big)\Big)$. According to Assumption~\ref{sup:iter_bound}, the inequality
	\begin{equation*}
		\begin{aligned}
			&\big\lvert \lambda^\mathcal{O}\big(\bfg_i,N_i(k+1)\big) - \lambda^\mathcal{O}\big(\bfg_i,N_i(k)\big) \big\lvert =  \\ 
			&\Bigg\lvert \sum_{e \in \Delta\mathcal{E}(\bfg_i,N_i,k+1)} \delta(e) \mD^{G^H_{ci}}(e)  \Bigg\lvert \leq 
			\Bigg\lvert \sum_{e \in \Delta\mathcal{E}(\bfg_i,N_i,k+1)} \mD^{G^H_{ci}}(e)  \Bigg\lvert \\
			&\leq \gamma \mathcal{B}
		\end{aligned}
	\end{equation*}
	holds for any iteration $k$ and robot $i \in V$.
	
	Next, according to the Step~2 to 4 in Appendix~A and the proof of Theorem~1, we can prove both Lemma~1 and Theorem~1 remain valid under the online exploration load metric \eqref{eq:online_load_metric}.
\end{proof}


\bibliography{IEEEabrv, sample}

\begin{thebibliography}{10}
\providecommand{\url}[1]{#1}
\csname url@samestyle\endcsname
\providecommand{\newblock}{\relax}
\providecommand{\bibinfo}[2]{#2}
\providecommand{\BIBentrySTDinterwordspacing}{\spaceskip=0pt\relax}
\providecommand{\BIBentryALTinterwordstretchfactor}{4}
\providecommand{\BIBentryALTinterwordspacing}{\spaceskip=\fontdimen2\font plus
\BIBentryALTinterwordstretchfactor\fontdimen3\font minus
  \fontdimen4\font\relax}
\providecommand{\BIBforeignlanguage}[2]{{%
\expandafter\ifx\csname l@#1\endcsname\relax
\typeout{** WARNING: IEEEtran.bst: No hyphenation pattern has been}%
\typeout{** loaded for the language `#1'. Using the pattern for}%
\typeout{** the default language instead.}%
\else
\language=\csname l@#1\endcsname
\fi
#2}}
\providecommand{\BIBdecl}{\relax}
\BIBdecl

\bibitem{2015_Bircher_ICRA}
A.~Bircher, K.~Alexis, M.~Burri, P.~Oettershagen, S.~Omari, T.~Mantel, and
  R.~Siegwart, ``Structural inspection path planning via iterative viewpoint
  resampling with application to aerial robotics,'' in \emph{Proc. IEEE Int.
  Conf. Robot. Autom.}, 2015, pp. 6423--6430.

\bibitem{2017_Erdlj}
M.~Erdelj, E.~Natalizio, K.~R. Chowdhury, and I.~F. Akyildiz, ``Help from the
  sky: Leveraging uavs for disaster management,'' \emph{IEEE Pervasive
  Comput.}, vol.~16, no.~1, pp. 24--32, 2017.

\bibitem{2012_Marconi}
L.~Marconi, C.~Melchiorri, M.~Beetz, D.~Pangercic, R.~Siegwart, S.~Leutenegger,
  R.~Carloni, S.~Stramigioli, H.~Bruyninckx, P.~Doherty, A.~Kleiner,
  V.~Lippiello, A.~Finzi, B.~Siciliano, A.~Sala, and N.~Tomatis, ``The sherpa
  project: Smart collaboration between humans and ground-aerial robots for
  improving rescuing activities in alpine environments,'' in \emph{IEEE Int.
  Symp. Saf., Secur., Rescue Robot.}, 2012, pp. 1--4.

\bibitem{Goal_assignment}
J.~Faigl, M.~Kulich, and L.~Přeučil, ``Goal assignment using distance cost in
  multi-robot exploration,'' in \emph{Proc. IEEE/RSJ Int. Conf. Intell. Robots
  Syst.}, 2012, pp. 3741--3746.

\bibitem{multi_dong_2019}
S.~Dong, K.~Xu, Q.~Zhou, A.~Tagliasacchi, S.~Xin, M.~Nie\ss{}ner, and B.~Chen,
  ``Multi-robot collaborative dense scene reconstruction,'' \emph{ACM Trans.
  Graph.}, vol.~38, no.~4, jul 2019.

\bibitem{RACER}
B.~Zhou, H.~Xu, and S.~Shen, ``Racer: Rapid collaborative exploration with a
  decentralized multi-uav system,'' \emph{{IEEE} Trans. Robot.}, vol.~39,
  no.~3, pp. 1816--1835, 2023.

\bibitem{pairwise}
L.~Klodt and V.~Willert, ``Equitable workload partitioning for multi-robot
  exploration through pairwise optimization,'' in \emph{Proc. IEEE/RSJ Int.
  Conf. Intell. Robots Syst.}, 2015, pp. 2809--2816.

\bibitem{bi_cure_2024}
Q.~Bi, X.~Zhang, J.~Wen, Z.~Pan, S.~Zhang, R.~Wang, and J.~Yuan, ``{CURE}: {A}
  {Hierarchical} {Framework} for {Multi}-{Robot} {Autonomous} {Exploration}
  {Inspired} by {Centroids} of {Unknown} {Regions},'' \emph{IEEE Trans. Autom.
  Sci. Eng.}, vol.~21, no.~3, pp. 3773--3786, Jul. 2024.

\bibitem{MR-TopoMap}
Z.~Zhang, J.~Yu, J.~Tang, Y.~Xu, and Y.~Wang, ``Mr-topomap: Multi-robot
  exploration based on topological map in communication restricted
  environment,'' \emph{IEEE Robot. Autom. Lett.}, vol.~7, no.~4, pp.
  10\,794--10\,801, 2022.

\bibitem{MR-DTG}
Q.~Dong, H.~Xi, S.~Zhang, Q.~Bi, T.~Li, Z.~Wang, and X.~Zhang, ``Fast and
  communication-efficient multi-uav exploration via voronoi partition on
  dynamic topological graph,'' in \emph{Proc. IEEE/RSJ Int. Conf. Intell.
  Robots Syst.}, 2024, pp. 14\,063--14\,070.

\bibitem{FALCON}
Y.~Zhang, X.~Chen, C.~Feng, B.~Zhou, and S.~Shen, ``Falcon: Fast autonomous
  aerial exploration using coverage path guidance,'' \emph{IEEE Trans. Robot.},
  vol.~41, pp. 1365--1385, 2025.

\bibitem{exploration_survey}
C.~Wang, C.~Yu, X.~Xu, Y.~Gao, X.~Yang, W.~Tang, S.~Yu, Y.~Chen, F.~Gao,
  Z.~Jian, X.~Chen, F.~Gao, B.~Zhou, and Y.~Wang, ``Multi-robot system for
  cooperative exploration in unknown environments: A survey,'' 2025,
  \textit{arXiv:2503.07278}.

\bibitem{zhou_fuel_2021}
B.~Zhou, Y.~Zhang, X.~Chen, and S.~Shen, ``{FUEL}: {Fast} {UAV} {Exploration}
  {Using} {Incremental} {Frontier} {Structure} and {Hierarchical} {Planning},''
  \emph{IEEE Robot. Autom. Lett.}, vol.~6, no.~2, pp. 779--786, Apr. 2021.

\bibitem{zhang_falcon_2025}
Y.~Zhang, X.~Chen, C.~Feng, B.~Zhou, and S.~Shen, ``{FALCON}: {Fast}
  {Autonomous} {Aerial} {Exploration} {Using} {Coverage} {Path} {Guidance},''
  \emph{IEEE Trans. Robot.}, vol.~41, pp. 1365--1385, 2025.

\bibitem{2024_huang_localization}
Y.~Huang, X.~Lin, and B.~Englot, ``Multi-robot autonomous exploration and
  mapping under localization uncertainty with expectation-maximization,'' in
  \emph{Proc. IEEE Int. Conf. Robot. Autom.}, 2024, pp. 7236--7242.

\bibitem{2024_tellaroli_iros}
M.~Tellaroli, M.~Luperto, M.~Antonazzi, and N.~Basilico, ``Frontier-based
  exploration for multi-robot rendezvous in communication-restricted unknown
  environments,'' in \emph{Proc. IEEE/RSJ Int. Conf. Intell. Robots Syst.},
  2024, pp. 5807--5812.

\bibitem{2023_papathrodorou_icra}
S.~Papatheodorou, N.~Funk, D.~Tzoumanikas, C.~Choi, B.~Xu, and S.~Leutenegger,
  ``Finding things in the unknown: Semantic object-centric exploration with an
  mav,'' in \emph{Proc. IEEE Int. Conf. Robot. Autom.}, 2023, pp. 3339--3345.

\bibitem{2024_luo_ral}
Y.~Luo, Z.~Zhuang, N.~Pan, C.~Feng, S.~Shen, F.~Gao, H.~Cheng, and B.~Zhou,
  ``Star-searcher: A complete and efficient aerial system for autonomous target
  search in complex unknown environments,'' \emph{IEEE Robot. Autom. Lett.},
  vol.~9, no.~5, pp. 4329--4336, 2024.

\bibitem{gonzalez-banos_navigation_2002}
H.~H. González-Baños and J.-C. Latombe, ``\BIBforeignlanguage{en}{Navigation
  {Strategies} for {Exploring} {Indoor} {Environments}},''
  \emph{\BIBforeignlanguage{en}{Int. J. Robot. Res.}}, vol.~21, no. 10-11, pp.
  829--848, Oct. 2002.

\bibitem{2016_bircher_icra}
A.~Bircher, M.~Kamel, K.~Alexis, H.~Oleynikova, and R.~Siegwart, ``Receding
  horizon "next-best-view" planner for 3d exploration,'' in \emph{Proc. IEEE
  Int. Conf. Robot. Autom.}, 2016, pp. 1462--1468.

\bibitem{2016_witting_iros}
C.~Witting, M.~Fehr, R.~Bähnemann, H.~Oleynikova, and R.~Siegwart,
  ``History-aware autonomous exploration in confined environments using mavs,''
  in \emph{Proc. IEEE/RSJ Int. Conf. Intell. Robots Syst.}, 2018, pp. 1--9.

\bibitem{1997_Yammauchi_CIRA}
B.~Yamauchi, ``A frontier-based approach for autonomous exploration,'' in
  \emph{Proc. IEEE Int. Symp. Comput. Intell. Robot. Automat. CIRA'97}, 1997,
  pp. 146--151.

\bibitem{2014_atanasov_cdc}
N.~Atanasov, R.~Tron, V.~M. Preciado, and G.~J. Pappas, ``Joint estimation and
  localization in sensor networks,'' in \emph{Proc. IEEE Conf. Decision
  Control}, 2014, pp. 6875--6882.

\bibitem{2024_ding_ral}
T.~Ding, R.~Zheng, S.~Zhang, and M.~Liu, ``Resource-efficient cooperative
  online scalar field mapping via distributed sparse gaussian process
  regression,'' \emph{IEEE Robot. Autom. Lett.}, vol.~9, no.~3, pp. 2295--2302,
  2024.

\bibitem{zobeidi_dense_2022}
E.~Zobeidi, A.~Koppel, and N.~Atanasov, ``Dense incremental metric-semantic
  mapping for multiagent systems via sparse gaussian process regression,''
  \emph{{IEEE} Trans. Robot.}, vol.~38, no.~5, pp. 3133--3153, Oct. 2022.

\bibitem{lajoie_swarm-slam_2024}
P.-Y. Lajoie and G.~Beltrame, ``Swarm-{SLAM}: {Sparse} {Decentralized}
  {Collaborative} {Simultaneous} {Localization} and {Mapping} {Framework} for
  {Multi}-{Robot} {Systems},'' \emph{IEEE Robot. Autom. Lett.}, vol.~9, no.~1,
  pp. 475--482, Jan. 2024.

\bibitem{2022_tian_tro}
Y.~Tian, Y.~Chang, F.~Herrera~Arias, C.~Nieto-Granda, J.~P. How, and
  L.~Carlone, ``Kimera-multi: Robust, distributed, dense metric-semantic slam
  for multi-robot systems,'' \emph{IEEE Trans. Robot.}, vol.~38, no.~4, pp.
  2022--2038, 2022.

\bibitem{ROAM}
A.~Asgharivaskasi, F.~Girke, and N.~Atanasov, ``Riemannian optimization for
  active mapping with robot teams,'' \emph{IEEE Trans. Robot.}, vol.~41, pp.
  1077--1097, 2025.

\bibitem{2025_ding_ral}
T.~Ding, R.~Zheng, S.~Zhang, and M.~Liu, ``Adaptive-resolution cooperative
  field mapping with event-triggered distributed map fusion,'' \emph{IEEE
  Robot. Autom. Lett.}, vol.~10, no.~2, pp. 1058--1065, 2025.

\bibitem{2022_dong_ral}
H.~Dong, J.~Yu, Y.~Xu, Z.~Xu, Z.~Shen, J.~Tang, Y.~Shen, and Y.~Wang,
  ``Mr-gmmapping: Communication efficient multi-robot mapping system via
  gaussian mixture model,'' \emph{IEEE Robot. Autom. Lett.}, vol.~7, no.~2, pp.
  3294--3301, 2022.

\bibitem{2023_Stathoulopoulos_icra}
N.~Stathoulopoulos, A.~Koval, A.-a. Agha-mohammadi, and G.~Nikolakopoulos,
  ``Frame: Fast and robust autonomous 3d point cloud map-merging for egocentric
  multi-robot exploration,'' in \emph{Proc. IEEE Int. Conf. Robot. Autom.},
  2023, pp. 3483--3489.

\bibitem{yang_active_2024}
X.~Yang, Y.~Yang, C.~Yu, J.~Chen, J.~Yu, H.~Ren, H.~Yang, and Y.~Wang, ``Active
  {Neural} {Topological} {Mapping} for {Multi}-{Agent} {Exploration},''
  \emph{IEEE Robot. Autom. Lett.}, vol.~9, no.~1, pp. 303--310, Jan. 2024.

\bibitem{2014_Gharesifard_tac}
B.~Gharesifard and J.~Cortés, ``Distributed continuous-time convex
  optimization on weight-balanced digraphs,'' \emph{IEEE Trans. Autom.
  Control}, vol.~59, no.~3, pp. 781--786, 2014.

\bibitem{2015_Nedic_tac}
A.~Nedić and A.~Olshevsky, ``Distributed optimization over time-varying
  directed graphs,'' \emph{IEEE Trans. Autom. Control}, vol.~60, no.~3, pp.
  601--615, 2015.

\bibitem{2014_geoffrey_ijrr}
G.~A. Hollinger and G.~S. Sukhatme, ``Sampling-based robotic information
  gathering algorithms,'' \emph{Int. J. Robot. Res.}, vol.~33, no.~9, pp.
  1271--1287, 2014.

\bibitem{2024_Jakkala_icra}
K.~Jakkala and S.~Akella, ``Multi-robot informative path planning from
  regression with sparse gaussian processes,'' in \emph{Proc. IEEE Int. Conf.
  Robot. Autom.}, 2024, pp. 12\,382--12\,388.

\bibitem{2024_Newaz_ral}
A.~A.~R. Newaz, P.~Padrao, J.~Fuentes, T.~Alam, G.~Govindarajan, and
  L.~Bobadilla, ``Lcd-rig: Limited communication decentralized robotic
  information gathering systems,'' \emph{IEEE Robot. Autom. Lett.}, vol.~9,
  no.~11, pp. 10\,034--10\,041, 2024.

\bibitem{2025_sun_tro}
M.~M. Sun, A.~Gaggar, P.~Trautman, and T.~Murphey, ``Fast ergodic search with
  kernel functions,'' \emph{IEEE Trans. Robot.}, vol.~41, pp. 1841--1860, 2025.

\bibitem{seewald_energy-aware_2024}
A.~Seewald, C.~J. Lerch, M.~Chancán, A.~M. Dollar, and I.~Abraham,
  ``Energy-{Aware} {Ergodic} {Search}: {Continuous} {Exploration} for
  {Multi}-{Agent} {Systems} with {Battery} {Constraints},'' in \emph{Proc. IEEE
  Int. Conf. Robot. Autom.}, May 2024, pp. 7048--7054.

\bibitem{BGE}
S.~Wu, C.~Wang, J.~Pan, D.~Han, and Z.~Zhao, ``Bayesian-guided evolutionary
  strategy with rrt for multi-robot exploration,'' in \emph{Proc. IEEE Int.
  Conf. Robot. Autom.}, 2024, pp. 12\,720--12\,726.

\bibitem{2012_Faigl_iros}
J.~Faigl, M.~Kulich, and L.~Přeučil, ``Goal assignment using distance cost in
  multi-robot exploration,'' in \emph{Proc. IEEE/RSJ Int. Conf. Intell. Robots
  Syst.}, 2012, pp. 3741--3746.

\bibitem{Voronoi_survey}
F.~Aurenhammer, ``\BIBforeignlanguage{English}{Voronoi diagrams - a survey of a
  fundamental geometric data structure},''
  \emph{\BIBforeignlanguage{English}{ACM Comput. Surv.}}, no.~23, pp. 345--405,
  1991.

\bibitem{clustering_survey}
R.~Xu and D.~Wunsch, ``Survey of clustering algorithms,'' \emph{IEEE Transa.
  Neural Netw.}, vol.~16, no.~3, pp. 645--678, 2005.

\bibitem{2010_Cortes_tac}
J.~Cortes, ``Coverage optimization and spatial load balancing by robotic sensor
  networks,'' \emph{IEEE Trans. Autom. Control}, vol.~55, no.~3, pp. 749--754,
  2010.

\bibitem{pavone_distributed_2011}
M.~Pavone, A.~Arsie, E.~Frazzoli, and F.~Bullo, ``Distributed {Algorithms} for
  {Environment} {Partitioning} in {Mobile} {Robotic} {Networks},'' \emph{IEEE
  Trans. Autom. Control}, vol.~56, no.~8, pp. 1834--1848, Aug. 2011.

\bibitem{boardman_limited_2017}
B.~Boardman, T.~Harden, and S.~Martínez, ``Limited range spatial load
  balancing for multiple robots,'' in \emph{Proc. Amer. Control Conf.}, May
  2017, pp. 2285--2290.

\bibitem{2024_zhang_acc}
H.~Zhang, R.~Zheng, S.~Zhang, and M.~Liu, ``A fully distributed, air-ground
  coordinated coverage control for multi-robot systems with limited sensing
  range,'' in \emph{Proc. Amer. Control Conf.}, 2024, pp. 1825--1830.

\bibitem{2016_Seoung_ijrr}
S.~K. Lee, S.~P. Fekete, and J.~McLurkin, ``Structured triangulation in
  multi-robot systems: Coverage, patrolling, voronoi partitions, and geodesic
  centers,'' \emph{Int. J. Robot. Res.}, vol.~35, no.~10, pp. 1234--1260, 2016.

\bibitem{2012_Durham_tro}
J.~W. Durham, R.~Carli, P.~Frasca, and F.~Bullo, ``Discrete partitioning and
  coverage control for gossiping robots,'' \emph{IEEE Trans. Robot.}, vol.~28,
  no.~2, pp. 364--378, 2012.

\bibitem{bhattacharya_multi-robot_2014}
S.~Bhattacharya, R.~Ghrist, and V.~Kumar, ``\BIBforeignlanguage{en}{Multi-robot
  coverage and exploration on {Riemannian} manifolds with boundaries},''
  \emph{\BIBforeignlanguage{en}{Int. J. Robot. Res.}}, vol.~33, no.~1, pp.
  113--137, Jan. 2014.

\bibitem{GraphVor}
M.~Erwig, ``The graph voronoi diagram with applications,'' \emph{Networks},
  vol.~36, no.~3, pp. 156--163, 2000.

\bibitem{2024_Bai_iros}
R.~Bai, S.~Yuan, H.~Guo, P.~Yin, W.-Y. Yau, and L.~Xie, ``Multi-robot active
  graph exploration with reduced pose-slam uncertainty via submodular
  optimization,'' in \emph{Proc. IEEE/RSJ Int. Conf. Intell. Robots Syst.},
  2024, pp. 10\,229--10\,236.

\bibitem{vutetakis_active_2024}
D.~Vutetakis and J.~Xiao, ``\BIBforeignlanguage{en}{Active perception network
  for non-myopic online exploration and visual surface coverage},''
  \emph{\BIBforeignlanguage{en}{Int. J. Robot. Res.}}, p. 02783649241264577,
  Aug. 2024.

\bibitem{de_groot_topology-driven_2025}
O.~de~Groot, L.~Ferranti, D.~M. Gavrila, and J.~Alonso-Mora,
  ``Topology-{Driven} {Parallel} {Trajectory} {Optimization} in {Dynamic}
  {Environments},'' \emph{IEEE Trans. Robot.}, vol.~41, pp. 110--126, 2025.

\bibitem{GVD_IJRR}
H.~C. Burdick;, ``Sensor-based exploration: The hierarchical generalized
  voronoi graph,'' \emph{Int. J. Robot. Res.}, vol.~19, no.~2, pp. 96--125,
  2000.

\bibitem{GVD_TRO}
J.~Y. Lee and H.~Choset, ``Sensor-based exploration for convex bodies: a new
  roadmap for a convex-shaped robot,'' \emph{IEEE Trans. Robot.}, vol.~21,
  no.~2, pp. 240--247, 2005.

\bibitem{2009_Nidhi_ras}
N.~Kalra, D.~Ferguson, and A.~Stentz, ``Incremental reconstruction of
  generalized voronoi diagrams on grids,'' \emph{Robot. Auton. Syst.}, vol.~57,
  no.~2, pp. 123--128, 2009.

\bibitem{wen_gvd_2025}
J.~Wen, X.~Zhang, Q.~Bi, H.~Liu, J.~Yuan, and Y.~Fang, ``G²{VD} {Planner}:
  {Efficient} {Motion} {Planning} {With} {Grid}-{Based} {Generalized} {Voronoi}
  {Diagrams},'' \emph{IEEE Trans. Autom. Sci. Eng.}, vol.~22, pp. 3743--3755,
  2025.

\bibitem{2019_zhou_ral}
B.~Zhou, F.~Gao, L.~Wang, C.~Liu, and S.~Shen, ``Robust and efficient quadrotor
  trajectory generation for fast autonomous flight,'' \emph{IEEE Robot. Autom.
  Lett.}, vol.~4, no.~4, pp. 3529--3536, 2019.

\bibitem{HELSGAUN2000106}
K.~Helsgaun, ``An effective implementation of the lin–kernighan traveling
  salesman heuristic,'' \emph{Eur. J. Oper. Res.}, vol. 126, no.~1, pp.
  106--130, 2000.

\end{thebibliography}
\bibliographystyle{IEEEtran}

\end{document}